\documentclass[10pt,twocolumn,letterpaper]{article}

\usepackage{cvpr}
\usepackage{times}
\usepackage{epsfig}
\usepackage{graphicx}
\usepackage{amsmath}
\usepackage{amssymb}

\usepackage[tableposition=above]{caption}
\usepackage{subcaption}
\usepackage{bbm}
\usepackage{bm}
\usepackage{xspace}
\usepackage[utf8]{inputenc} 
\usepackage{url}            
\usepackage{booktabs}       
\usepackage{amsfonts}       
\usepackage{nicefrac}       
\usepackage{microtype}      
\usepackage{color}
\usepackage{algorithm}
\usepackage{algorithmic}
\usepackage{array}
\usepackage{multirow}
\usepackage{enumitem}
\usepackage{makecell}
\usepackage{gensymb}         
\usepackage{mathtools}
\usepackage{dblfloatfix}

\makeatletter
\DeclareRobustCommand\onedot{\futurelet\@let@token\@onedot}
\def\@onedot{\ifx\@let@token.\else.\null\fi\xspace}

\def\eg{\emph{e.g}\onedot}

\def\etal{\emph{et al}\onedot}
\makeatother

\usepackage{xcolor}
\usepackage{ifthen}
\usepackage{textcomp}
\definecolor{red}{rgb}{0.9,0.1,0}
\definecolor{slateblue}{rgb}{0.7,0.35,0.9}
\definecolor{green}{rgb}{0, 0.4, 0}
\definecolor{brown}{rgb}{0.3, 0.2, 0}
\definecolor{mahogany}{rgb}{0.75, 0.25, 0.0}
\definecolor{purple}{rgb}{0.6, 0, 0.6}
\definecolor{darkpurple}{rgb}{0.3, 0, 0.3}
\definecolor{darkgreen}{rgb}{0, 0.4, 0}
\definecolor{frenchblue}{rgb}{0.0, 0.45, 0.73}
\definecolor{blue}{rgb}{0.0, 0.0, 1.0}
\definecolor{goldenrod}{rgb}{0.65, 0.45, 0.03}
\definecolor{gray}{rgb}{0.5,0.5,0.5}

\newboolean{revising}
\setboolean{revising}{true}
\ifthenelse{\boolean{revising}}
{
    \newcommand{\ignore}[1]{}

} {
    \newcommand{\ignore}[1]{}

}

\makeatletter
\renewcommand{\paragraph}{%
  \@startsection{paragraph}{4}%
  {\z@}{0.5\baselineskip \@plus 0ex \@minus 0ex}{-1em}%
  {\normalfont\normalsize\bfseries}%
}
\makeatother

\makeatletter
\newcommand\footnoteref[1]{\protected@xdef\@thefnmark{\ref{#1}}\@footnotemark}
\makeatother

\usepackage[pagebackref=true,breaklinks=true,letterpaper=true,colorlinks,bookmarks=false]{hyperref}

\cvprfinalcopy 

\def\cvprPaperID{983} 

\makeatletter
\let\@fnsymbol\@arabic
\makeatother
\begin{document}

\title{HoHoNet: 360 Indoor Holistic Understanding with Latent Horizontal Features}


\author{
Cheng Sun\thanks{National Tsing Hua University}~$^{,}$\thanks{ASUS AICS Department} \\ \href{mailto:chengsun@gapp.nthu.edu.tw}{\small\tt \textcolor{black}{chengsun@gapp.nthu.edu.tw}}
\and
Min Sun\footnotemark[1]~$^{,}$\thanks{Joint Research Center for AI Technology and All Vista Healthcare} \\ \href{mailto:sunmin@ee.nthu.edu.tw}{\small\tt \textcolor{black}{sunmin@ee.nthu.edu.tw}}
\and
Hwann-Tzong Chen\footnotemark[1]~$^{,}$\thanks{Aeolus Robotics} \\ \href{mailto:htchen@cs.nthu.edu.tw}{\small\tt \textcolor{black}{htchen@cs.nthu.edu.tw}}
}

\maketitle
\thispagestyle{empty}

\begin{abstract}
    We present HoHoNet, a versatile and efficient framework for holistic understanding of an indoor 360-degree panorama using a Latent Horizontal Feature (LHFeat).
    The compact LHFeat flattens the features along the vertical direction and has shown success in modeling per-column modality for room layout reconstruction.
    HoHoNet advances in two important aspects. First, the deep architecture is redesigned to run faster with improved accuracy.
    Second, we propose a novel horizon-to-dense module, which relaxes the per-column output shape constraint, allowing per-pixel dense prediction from LHFeat.
    HoHoNet is fast: It runs at 52 FPS and 110 FPS with ResNet-50 and ResNet-34 backbones respectively, for modeling dense modalities from a high-resolution $512 \times 1024$ panorama. HoHoNet is also accurate. On the tasks of layout estimation and semantic segmentation, HoHoNet achieves results on par with current state-of-the-art. On dense depth estimation, HoHoNet outperforms all the prior arts by a large margin.
    Code is available at \url{https://github.com/sunset1995/HoHoNet}.
\end{abstract}

\section{Introduction}

Panoramic images can capture the complete $360\degree$ FOVs in one shot to provide a wide range of context that facilitates scene understanding~\cite{ZhangSTX14}.
As omnidirectional cameras become more easily accessible and several large-scale panorama datasets have been released, a growing number of techniques are developed for tasks of panoramic scene modeling such as semantic segmentation~\cite{EderSLF20,LeeJYCY19,ZhangLSC19}, depth estimation~\cite{JinXZZTXYG20,WangYSCT20,ZengKG20}, layout reconstruction~\cite{SunHSC19,YangWPWSC19,ZouSPCSWCH19}, and indoor real-time navigation~\cite{ChaplotSGG20}.

\begin{figure}
    \centering
    \includegraphics[width=\linewidth]{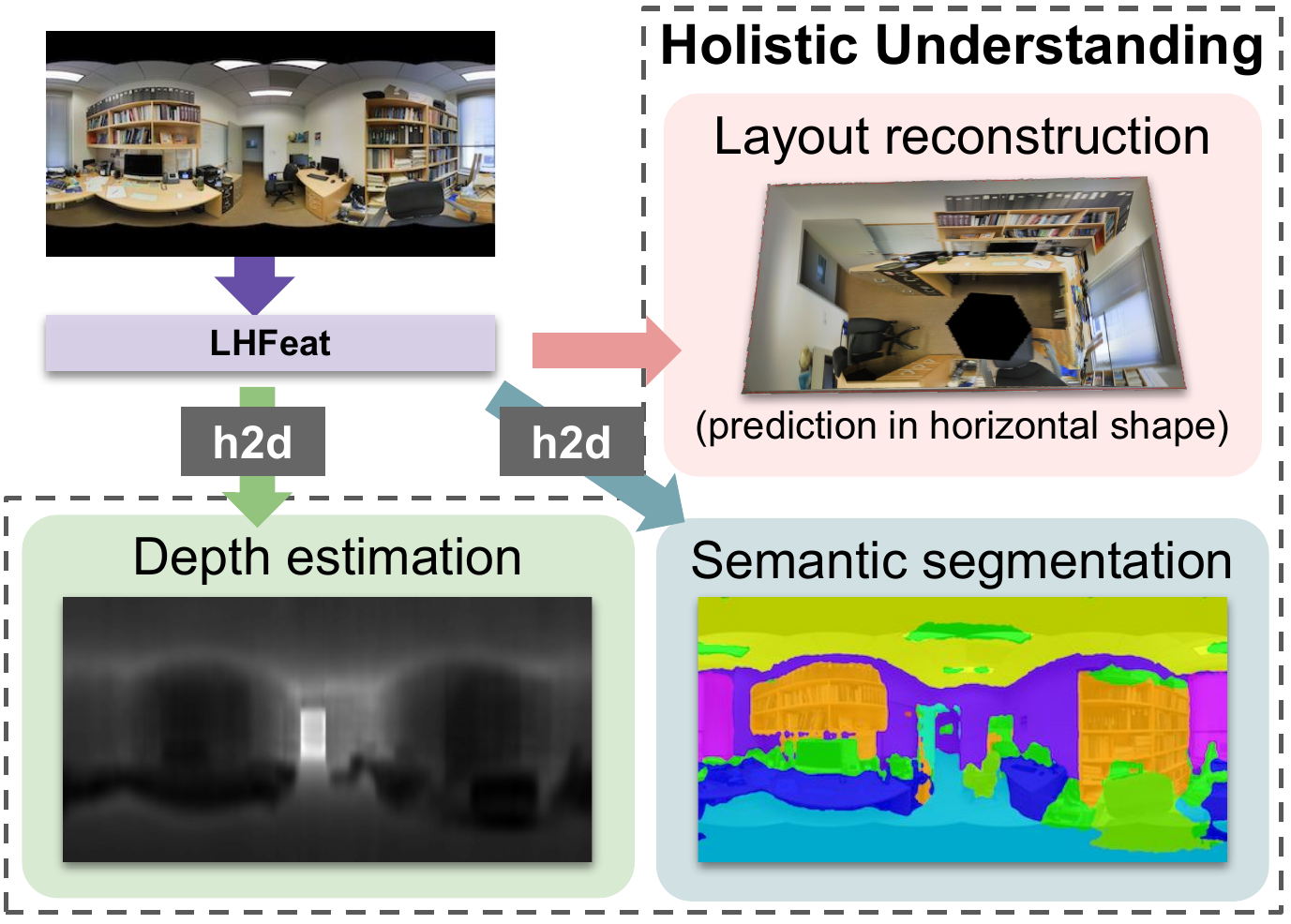}
    \caption{
       One framework for all: HoHoNet is a novel deep learning framework for modeling layout structure, dense depth, and semantic segmentation through a \emph{Latent Horizontal Feature representation} (LHFeat) whose height dimension is flattened.
       The proposed horizon-to-dense (h2d) module can produce dense predictions from the compact LHFeat.
    }
    \label{fig:teaser}
    \vspace{-1em}
\end{figure}

This paper aims to address the problem of holistic scene modeling from a single high-resolution equirectangular projection (ERP) image that captures the $360\degree$ panorama. We present HoHoNet as an efficient, effective, and versatile framework to achieve this goal (Fig.~\ref{fig:teaser}). 
The input ERP image is first passed through a CNN backbone for feature pyramid extraction, and then a proposed efficient height compression module encodes the feature pyramid into a \emph{Latent Horizontal Feature representation} (LHFeat) whose height dimension is flattened.
Finally, from LHFeat, the HoHoNet framework can yield both \emph{per-column} and \emph{per-pixel} modalities with state-of-the-art quality.

\begin{figure}
    \centering
    \begin{subfigure}[t]{0.325\linewidth}
        \centering
        \includegraphics[width=\linewidth]{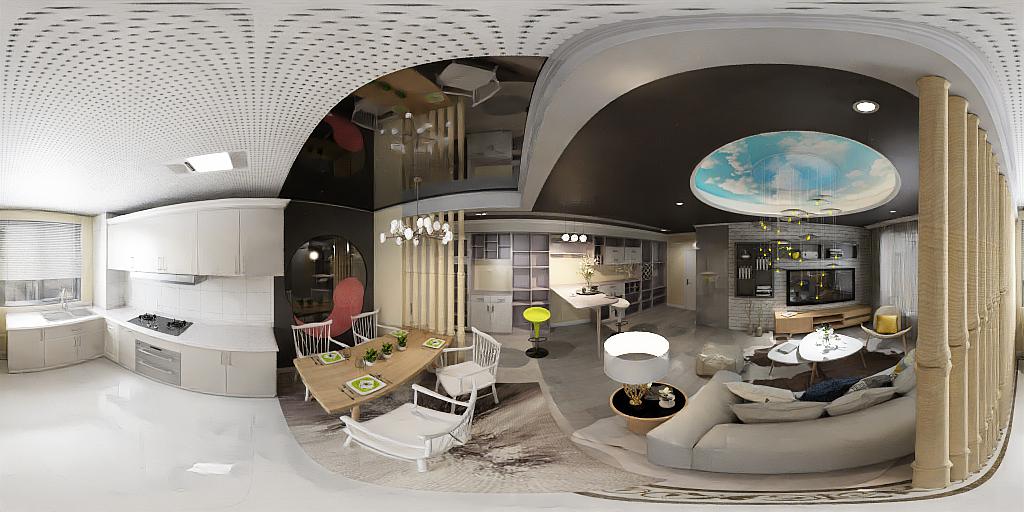}
        \caption{Aligned 360.}
    \end{subfigure}
    \hfill
    \begin{subfigure}[t]{0.325\linewidth}
        \centering
        \includegraphics[width=\linewidth]{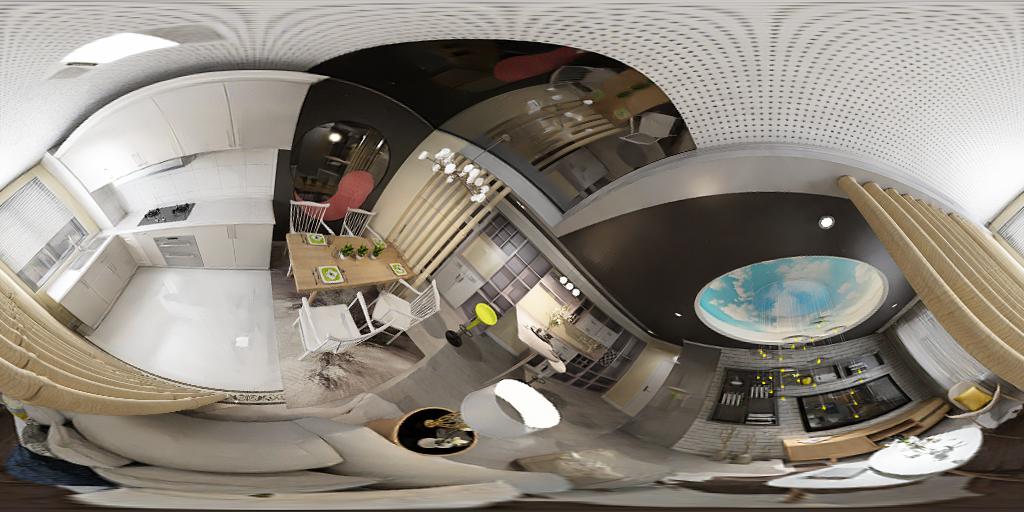}
        \caption{Roll rotation.}
        \label{fig:roll_rotation}
    \end{subfigure}
    \hfill
    \begin{subfigure}[t]{0.325\linewidth}
        \centering
        \includegraphics[width=\linewidth]{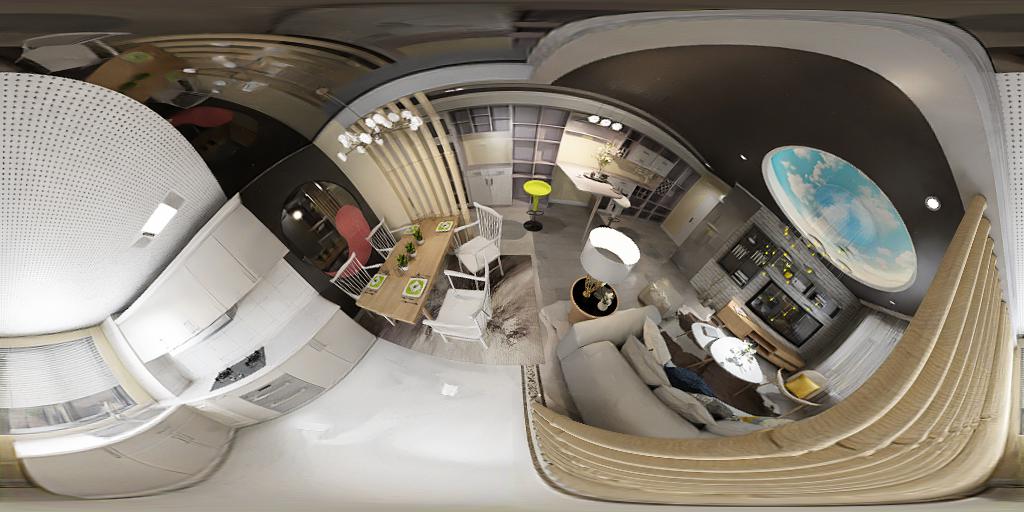}
        \caption{Pitch rotation.}
        \label{fig:pitch_rotation}
    \end{subfigure}
    \par\medskip
    \begin{subfigure}[t]{0.99\linewidth}
        \centering
        \includegraphics[width=0.9\linewidth]{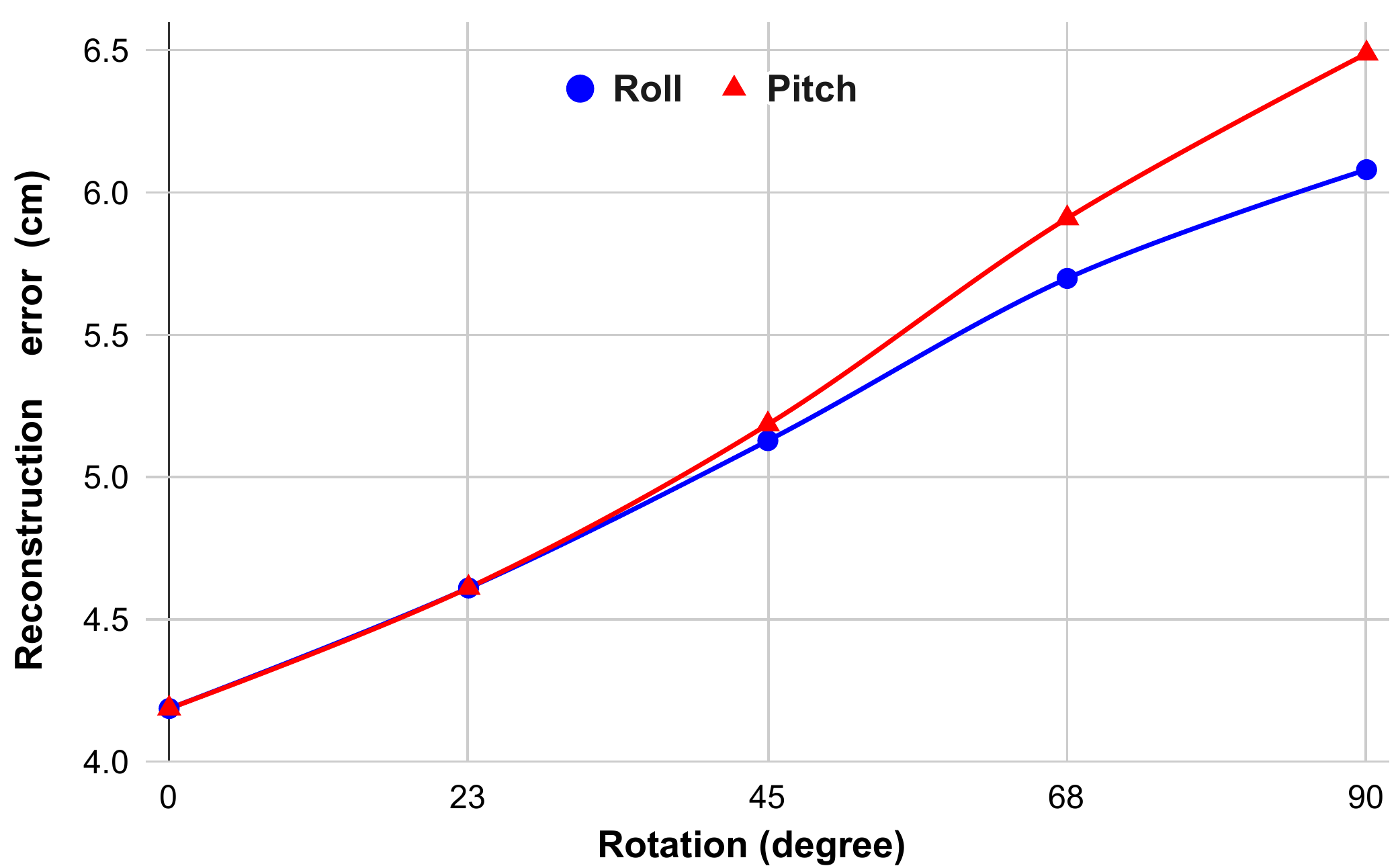}
        \caption{
        Gravity-aligned 360 image columns are easier to compress. 
        }
    \end{subfigure}

    \caption{
        We show that the structure information of an image column can be better kept in compression when the $y$-axis of the image is gravity aligned.
        We sample 1000 depth maps from Structured3D~\cite{ZhengZLTGZ20} dataset for the statistic.
        A $512 \times 1024$ depth map is compressed to $16 \times 1024$ via discrete cosine transform with high frequency truncated, which is applied to each column separately.
        We measure the absolute error between the original depth and the inverse transformed one.
    }
    \label{fig:depth_motivation}
    \vspace{-1em}
\end{figure}

Our way of encoding ERP images into LHFeat is inspired by Sun~\etal~\cite{SunHSC19}.
However, their model is only applicable to tasks of predicting per-column modalities (\eg, corners or boundaries of layout), which constrains its feasibility in other scenarios requiring per-pixel predictions.
We show that LHFeat can flexibly encode latent features for recovering the target 2D per-pixel modalities, based on our observation of the strong regularity between human-made structures and gravity aligned $y$-axis of ERP images (Fig.~\ref{fig:depth_motivation}).

In HoHoNet we introduce a new horizon-to-dense (h2d) module for recovering 2D per-pixel modalities while maintaining the efficiency of overall framework (Fig.~\ref{fig:teaser}).
A naive method is to treat the channel dimension of horizontal prediction as height and apply a linear interpolation if required.
However, this requires the shallow $\operatorname{Conv1D}$ layers to disentangle the row-dependent information from the row-independent LHFeat.
The spatial (the row) blended essence of LHFeat motivates us to model dense information in the frequency domain, and we resort to the discrete cosine transform (DCT) for its long-standing applications in data compression.
By replacing linear interpolation with IDCT, we are able to improve the dense prediction results.
With our horizon-to-dense module, the efficiently encoded LHFeat can now model dense modalities.

We summarize the key merits and contributions of HoHoNet for holistic scene modeling from a $360\degree$ image.
\begin{itemize}
    \setlength\itemsep{0em}
    \item \textbf{Fast.}
        HoHoNet can yield dense modalities for a high-resolution $512 \times 1024$ panorama at 52 FPS and 110 FPS with ResNet-50 and ResNet-34 respectively.
    \item \textbf{Versatile.}
        Our method relaxes the final prediction space upon the compact LHFeat from $\mathcal{O}(W)$ to the most common $\mathcal{O}(HW)$, capable of modeling layout, dense depth, and semantic segmentation.
    \item \textbf{Accurate.}
        The performances of HoHoNet on semantic segmentation and layout reconstruction are on par with the recent state-of-the-art.
        On dense depth estimation, HoHoNet outperforms prior arts by a margin.
\end{itemize}

\section{Related work}

\paragraph{Indoor 360 datasets.}
Scene modeling on $360\degree$ images is a topic with a growing number of researches recently.
Several 360 datasets are released to facilitate the learning-based methods.
Stanford2D3D~\cite{ArmeniSZS17} and Matterport3D~\cite{ChangDFHNSSZZ17} datasets are currently the two largest real-world indoor 360 datasets with various modalities being provided.
To model the higher level indoor structure, human-annotated layout datasets~\cite{WangYSCT20Layout,YangWPWSC19,ZhangSTX14,ZouCSH18,ZouSPCSWCH19} are constructed with more data and topology.
Structured3D~\cite{ZhengZLTGZ20} is a recently published photorealistic 360 dataset with abundant data and structure annotations from virtual environments.
In this work, we focus on real-world datasets to model depth, semantic, and layout modalities.

\begin{figure*}
    \centering
    \includegraphics[width=0.9\linewidth]{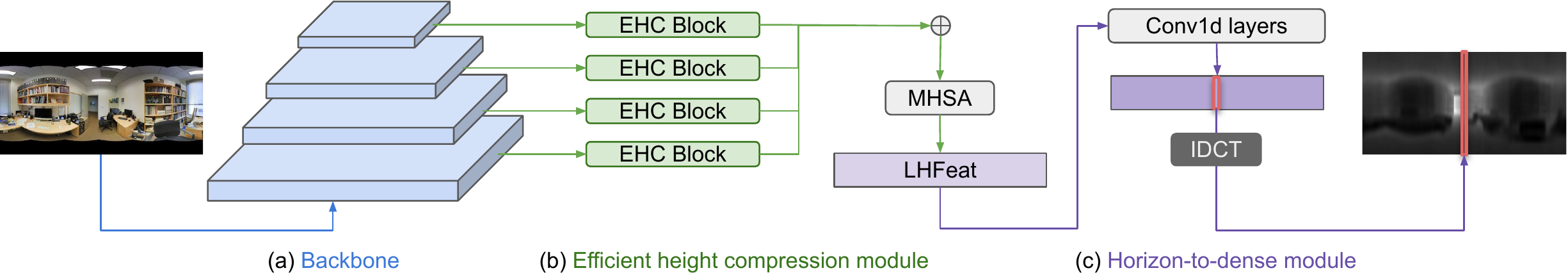}
    \caption{
        An overview of the HoHoNet framework for dense depth estimation.
        (a) A high-resolution panorama is first processed by the backbone (\eg, ResNet).
        (b) The feature pyramid is then squeezed and fused by the proposed Efficient Height Compression (EHC) module, with a Multi-Head Self-Attention (MHSA) for refinement (detailed in Sec.~\ref{ssec:horizontal_feat}).
        Note that the resulting LHFeat is compact (\eg, it is  $\mathbb{R}^{256 \times 1024}$ if the input image is $\mathbb{R}^{3 \times 512 \times 1024}$), enabling the overall network to run much faster than conventional encoder-decoder networks for dense features.
        (c) Finally, 1D convolution layers are employed to yield the final prediction.
        We find that predicting in DCT frequency domain brings about superior results, so we apply IDCT to the prediction of each column (detailed in Sec.~\ref{ssec:pred_2d}).
        Sec.~\ref{sec:approach} and supplementary material contain more architectural details.
    }
    \label{fig:overview}
    \vspace{-1em}
\end{figure*}

\paragraph{Input 360 format.}
Three standard 360 input formats are commonly used in the literature---{\it i)} equirectangular projection (ERP), {\it ii)} multiple perspective projections, and {\it iii)} icosahedron mesh.
ERP preserves all captured information in one image, but it also introduces distortion that might degrade the performance of the conventional convolution layer designed for perspective imagery.
A number of variants of convolution layers~\cite{CohenGKW18,CoorsCG18,SuG17,SuG19,TatenoNT18} have been proposed to address the issue of ERP distortion.
Projecting the $360\degree$ signal to multiple planar images makes it applicable to use classical CNNs with plenty of pre-trained models available, but the FOV of each view is limited.
Several padding~\cite{ChengCDWLS18,WangYSCT20} and view sampling~\cite{EderSLF20} strategies are proposed to deliver context information between views.
Recently, a few approaches propose to represent the omnidirectional input via icosahedron mesh for scene modeling~\cite{LeeJYCY19,ZhangLSC19}.
In this work, our model takes ERP as the $360\degree$ input format and apply classical convolution layers directly.
Although we speculate that incorporating distortion-aware techniques into our model with extra computational overheads could potentially improve performance, for the sake of simplicity and efficiency, we do not digress to pursue in that direction as the proposed method already achieves state-of-the-art performance.

\paragraph{Depth estimation on 360 imagery.}
To model depth on omnidirectional imagery, OmniDepth~\cite{ZioulisKZD18} designs encoder-decoder architectures considering the ERP distortion.
PanoPopups~\cite{EderMG19} shows that learning 360 depth with plane-aware loss is beneficial in the synthetic environment.
Recent works on panorama dense depth estimation propose to jointly learn from different projections~\cite{WangYSCT20} or different modalities~\cite{JinXZZTXYG20,ZengKG20}.
In contrast to most recent methods~\cite{JinXZZTXYG20,WangYSCT20,ZengKG20} that employ multiple backbones with cascaded training stages, HoHoNet consists of only one backbone and is trained in only one stage.
Besides, HoHoNet models dense depth through the  compact LHFeat while the prior arts estimate depth from conventional dense features.

\paragraph{Semantic segmentation on 360 imagery.}
Semantic segmentation is a fundamental task for scene modeling.
DistConv~\cite{TatenoNT18} proposes a distortion-aware deformable convolution layer for dense depth and semantic prediction on ERP images.
Most of the recent methods for 360 semantic segmentation design a trainable layer operating on representation related to icosahedral mesh~\cite{CohenWKW19,JiangHKPMN19,LeeJYCY19,ZhangLSC19}.
However, all methods above run on a relatively low resolution for the panoramic signal.
Tangent images~\cite{EderSLF20} project omnidirectional signals to multiple planar images tangent to a subdivided icosahedron, which allows to process high-resolution panoramas and to deploy the pre-trained weights on perspective images.
Similar to~\cite{EderSLF20}, HoHoNet can also operate on a high-resolution image, which is shown to be an essential factor in achieving better semantic segmentation accuracy.
In contrast to the recent methods, HoHoNet runs on ERP images directly, and the highly optimized deep-learning library can easily implement all our operations.

\paragraph{Latent horizontal features (LHFeat).}
HoHoNet is closely related to HorizonNet~\cite{SunHSC19} on the motivation of using 1D features. However, HorizonNet only tackles a specific layout reconstruction task and can only predict horizontal modalities.
We design a new architecture for encoding the LHFeat with much better speed and accuracy, and, importantly, we relax the constraint on output space via the proposed \emph{horizon-to-dense} module, which enables dense-modality holistic scene modeling.
We show that the compact LHFeat can be effectively applied to more tasks including dense depth estimation and semantic segmentation.

\section{Approach} \label{sec:approach}

\subsection{Framework overview} \label{ssec:overview}

An overview of the proposed framework is depicted in Fig.~\ref{fig:overview}. We describe the details below.

\paragraph{Input 360 image.}
We use the standard equirectangular projection (ERP) for $360\degree$ images. 
The resolution of input ERP images, $H_{\mathrm{inp.}} \times W_{\mathrm{inp.}}$, is a hyperparameter, and we set it according to the standard practice of each benchmark.
We show in Fig.~\ref{fig:depth_motivation} that the structure signals of an image column are preserved better after compression if the gravity direction is aligned with the image's $y$-axis, which is also a desirable property for our framework to encode a column into a latent vector.
In this work, the 360 data provided by the benchmarks are mostly well-aligned, so we do not apply any pre-processing.
Future applications could consider using the IMU sensor or 360 VP detection algorithm~\cite{ZhangSTX14,ZouCSH18} to pre-process and align the input for better robustness.

\paragraph{Backbone.}
We adopt ResNet~\cite{HeZRS16}, and the intermediate features from the four ResNet stages form the feature pyramid---$\{\mathbb{R}^{C_\ell \times H_\ell \times W_\ell}\}_{\ell=1,2,3,4}$ where $H_\ell = \frac{H_{\mathrm{inp.}}}{2^{\ell+1}}$, $W_\ell = \frac{W_{\mathrm{inp.}}}{2^{\ell+1}}$ and $C_\ell$ is the latent dimension of ResNet.

\paragraph{Extracting latent horizontal features (LHFeat).}
We propose an efficient height compression (EHC) module to extract the LHFeat $\mathbb{R}^{D \times W_1}$ from the backbone's feature pyramid.
We detail the EHC module in Sec.~\ref{ssec:horizontal_feat}.

\paragraph{Predicting modalities.}
We use $N$ in this work to denote the number of target channels for a task (\eg, $N$ is set to $1$ for depth estimation and is set to the number of classes for semantic segmentation).
Given the LHFeat $\mathbb{R}^{D \times W_1}$, we show how HoHoNet predicts 1D output $\mathbb{R}^{N \times W_{\mathrm{inp.}}}$ in Sec.~\ref{ssec:pred_1d}.
In Sec.~\ref{ssec:pred_2d}, we propose the first method to yield 2D dense prediction $\mathbb{R}^{N \times H_{\mathrm{inp.}} \times W_{\mathrm{inp.}}}$ from the compact LHFeat, which widely extends the potential applications of the proposed efficient framework.

\begin{figure}
    \centering
    \includegraphics[width=.9\linewidth]{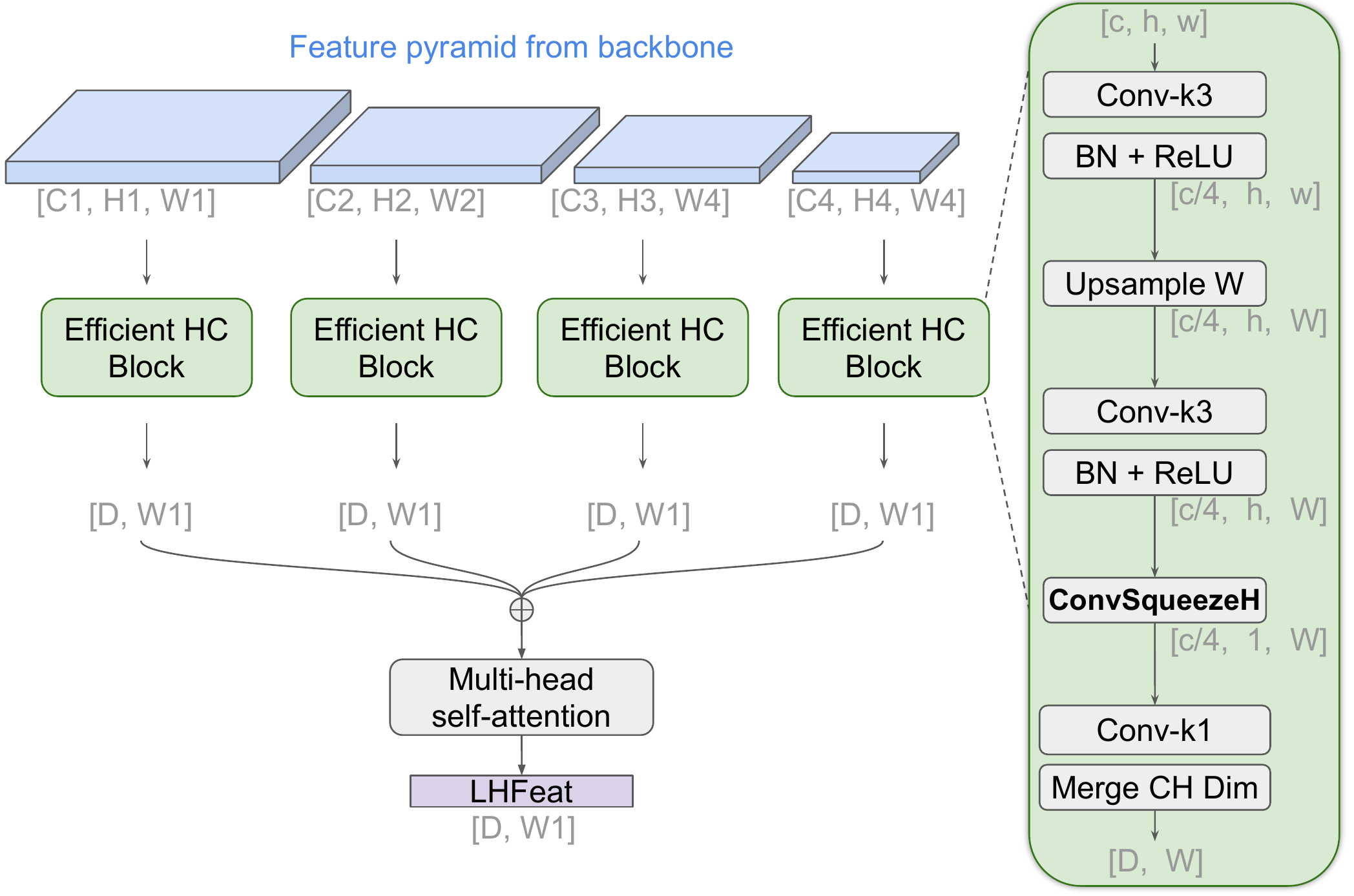}
    \caption{
        The proposed efficient height compression (EHC) module.
        The sizes of 2D and 1D features are denoted as $[C, H, W]$ and $[C, W]$ respectively.
        The $\operatorname{ConvSqueezeH}$ layer is a depthwise convolution layer with kernel size set to the prior known input feature height without padding, which produces output feature height $1$.
        See Sec.~\ref{ssec:horizontal_feat} for details.
    }
    \label{fig:efficient_hc}
    \vspace{-1em}
\end{figure}

\subsection{EHC module for LHFeat} \label{ssec:horizontal_feat}
The proposed efficient height compression (EHC) module is illustrated in Fig.~\ref{fig:efficient_hc}.
We first employ EHC blocks to squeeze the height of each 2D feature from the backbone's pyramid.
The resulting 1D features are then simply fused by summation.
Within the EHC block, the input 2D features are first processed by a $\operatorname{Conv2D}$ block for channel reduction, and then the spatial width is upsampled to $W_1$ if needed, and finally, another $\operatorname{Conv2D}$ block refines the upsampled features.
To efficiently reduce the feature height to $1$, we design the $\operatorname{ConvSqueezeH}$ layer, a depthwise convolution layer with kernel size set to $(h, 1)$ to cover full feature height without padding.
Note that the parameter $h$ of each EHC block is automatically pre-computed given $H_{\mathrm{inp.}}$.
Finally, a $\operatorname{Conv2D}$ layer converts the number of channels to LHFeat's latent size $D$, and the height dimension is simply discarded as it is already reduced to $1$ by the $\operatorname{ConvSqueezeH}$ layer.

To further refine the initial LHFeat, the similar prior work~\cite{SunHSC19} adopts bidirectional LSTM~\cite{HochreiterS97} for horizontal prediction.
We find the recurrent layer accounts for 22\% of our deep net processing time, so we employ multi-head self-attention~\cite{VaswaniSPUJGKP17} (MHSA) instead.
Our results show that MHSA runs faster and improves accuracy more.

\subsection{Predicting 1D per-column modalities} \label{ssec:pred_1d}
The target modality of some applications can be formulated into per-column prediction instead of the conventional per-pixel format. An example in this regard has been shown by Sun~\etal~\cite{SunHSC19} for layout estimation.
To predict the 1D modalities, we first upsample the horizontal features from $\mathbb{R}^{D \times W_1}$ to $\mathbb{R}^{D \times W_{\mathrm{inp.}}}$ and apply three $\operatorname{Conv1D}$ layers of kernel size $3$, $3$, and $1$ respectively with $\operatorname{BN}$, $\operatorname{ReLU}$ in between.
The last layer yields the final prediction in $\mathbb{R}^{N \times W_{\mathrm{inp.}}}$.

\begin{figure}
    \centering
    \includegraphics[width=1.0\linewidth]{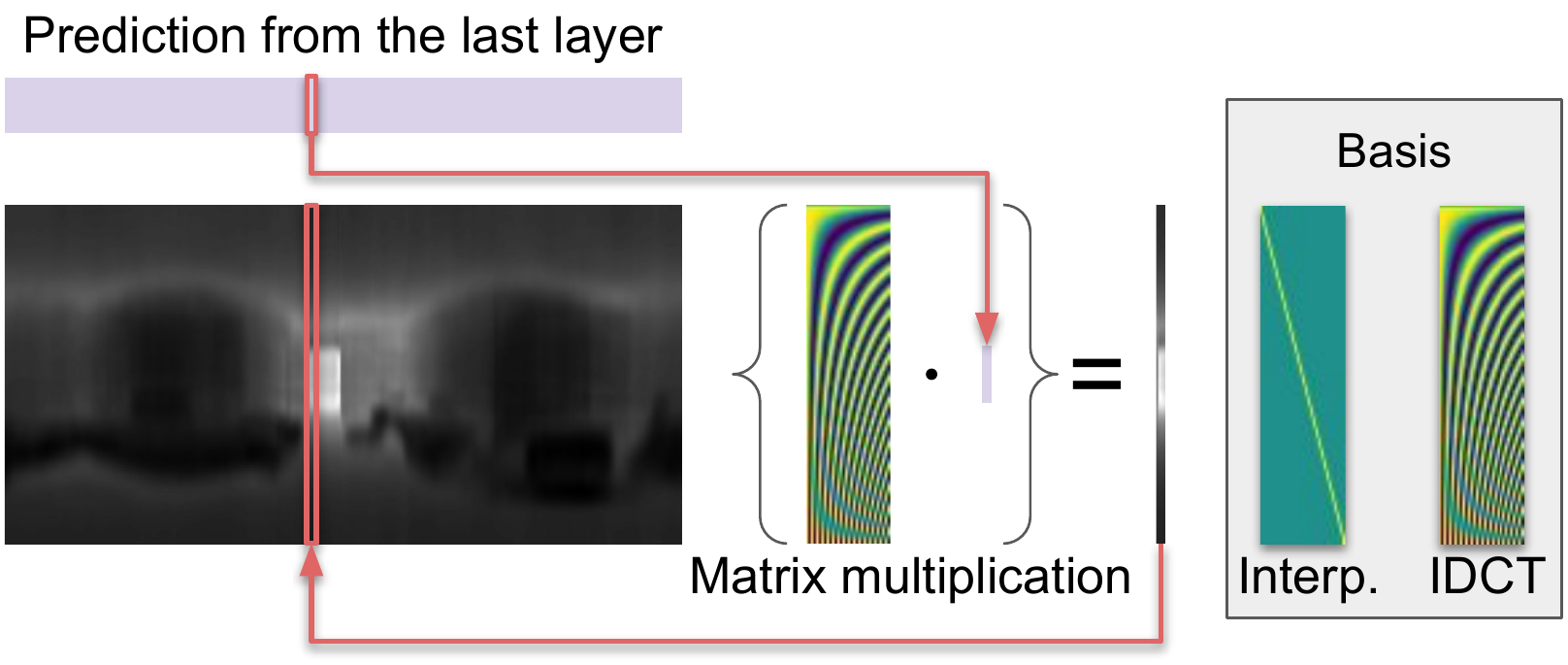}
    \caption{
        The predictions at each column act as the weights for the linear combination of components in basis $M$.
        HoHoNet learns to predict in the spatial domain if $M$ implements linear interpolation, and learns in the frequency domain if $M$ implements IDCT.
        See Sec.~\ref{ssec:pred_2d} for details.
    }
    \label{fig:unified_view}
    \vspace{-1em}
\end{figure}

\subsection{Predicting 2D per-pixel modalities} \label{ssec:pred_2d}
The strategy of shaping output space into per-column format does not apply to tasks that involve per-pixel modalities.
Here we present the horizon-to-dense module of HoHoNet to derive dense prediction $\mathbb{R}^{N \times H_{\mathrm{inp.}} \times W_{\mathrm{inp.}}}$ from the compact LHFeat $\mathbb{R}^{D \times W_1}$. This functionality opens the door to a more common scenario for various applications.

The trainable layers for 2D modality prediction are almost the same as the layers for 1D prediction introduced in Sec.~\ref{ssec:pred_1d} except that the number of channels in the output layer is augmented to $E = N\cdot r$ where $N$ is the number of target channels for a task and $r$ is the number of components shared by a image column.
The produced prediction is then reshaped from $\mathbb{R}^{E \times W_{\mathrm{inp.}}}$ to $\mathbb{R}^{N \times r \times W_{\mathrm{inp.}}}$.
We present two different operations to recover $\mathbb{R}^r$ back to $\mathbb{R}^{H_{\mathrm{inp.}}}$ for each column depending on the physical meaning we assign to the $r$ predicted values.

\paragraph{Interpolation.}
The simplest way is to view the latent dimension $r$ as the output height and apply linear interpolation to resize $r$ to $H_{\mathrm{inp.}}$ if $r < H_{\mathrm{inp.}}$.

\paragraph{Inverse discrete cosine transform (IDCT).}
Inspired by the application of the DCT in image compression for its energy compaction property, we view the $r$ predicted values as if they are in the DCT frequency domain with higher frequencies being truncated.
In this case, we can apply IDCT to recover the low-pass signal back to the original signal.
Let $x = [x_n]_{n=0}^{r-1} \in \mathbb{R}^r$ be the prediction; the final output $X = [X_m]_{m=0}^{H-1} \in \mathbb{R}^H$ can be recovered by
\begin{equation}
    X_m = \frac{x_0}{2} + \sum_{n=1}^{r-1} x_n \cos\left[
        \frac{\pi}{H}n\left(m + \frac{1}{2}\right)
    \right].
\end{equation}

\paragraph{A unified view.}
We can put the two aforementioned operations into a unified view of matrix multiplication as $X = Mx$ where $x \in \mathbb{R}^{r}$, $X \in \mathbb{R}^{H}$, and $M \in \mathbb{R}^{H \times r}$ consisting of $r$ orthogonal column vectors. Depending on the choice of basis, this unified view can implement linear interpolation or IDCT, as shown in Fig.~\ref{fig:unified_view}. 

We find that IDCT constantly outperforms linear interpolation.
We elaborate the intuition as follows.
The LHFeat blends the spatial-row information (as described in Sec.~\ref{ssec:horizontal_feat}), so training the last layers to disentangle the row-dependent dense modality from the flattened row-less LHFeat would pose a challenge.
Conversely, learning to predict in the frequency domain can benefit from the well defined basis functions with meaningful spatial frequencies that characterize each column's original row information as a whole, and therefore may alleviate the row-dependency problem.

\section{Experiments}
In Sec.~\ref{ssec:exp_ablation}, we first conduct ablation studies for the proposed components in HoHoNet.
We then compare the performance of HoHoNet with state-of-the-art methods on dense depth estimation (Sec.~\ref{ssec:exp_depth}), semantic segmentation (Sec.~\ref{ssec:exp_semantic}), and layout estimation (Sec.~\ref{ssec:exp_layout}).
Note that we train HoHoNet for each task separately and focus on showcasing the effectiveness of HoHoNet in learning a modality.
In Sec.~\ref{ssec:poor_pose}, we analyze the effect of non-gravity-aligned view.
More quantitative and qualitative results are included in the supplementary material.

\subsection{Ablation study} \label{ssec:exp_ablation}

Table~\ref{tab:abla_h2d} summarizes the results of ablation experiments, where we compare different settings of HoHoNet for dense depth estimation. Detailed descriptions are as follows.

\paragraph{\emph{Ablation split} for Matterport3D~\cite{ChangDFHNSSZZ17}.}
Matterport3D is a large-scale real-world dataset of indoor panoramas.
We prepare the \emph{ablation split} by splitting the official 61 training houses into 41 and 20 houses (containing 4{,}921 and 2{,}908 panoramas) for training and validation during ablation study.
We do not use the official validation split for ablation study as it will be used for state-of-the-art comparison later.
The input ERPs are resized to $512 \times 1024$.

\paragraph{Training and evaluation.}
We use Adam~\cite{KingmaB14} to optimize the L1 loss for 40 epochs with batch-size of 4.
The learning rate is set to 1e-4, and we apply polynomial learning rate decay with factor $0.9$.
Standard depth evaluation metric---MAE, RMSE, and $\delta^1$---are used.
We measure the average frame per second (FPS) for processing 50 individual $512 \times 1024$ panoramas on a GeForce RTX 2080 Ti.

\paragraph{Architecture of LHFeat extraction.}
Table~\ref{tab:abla_ehc} compares the proposed efficient height compression (EHC) module with the architecture used in the related work~\cite{SunHSC19}.
In~\cite{SunHSC19}, a sequence of convolution layers gradually reduces the feature heights to form the initial LHFeat, which is then followed by a bidirectional LSTM (Bi-LSTM) for feature refinement. (Detailed architectures are in the supplementary material.)
Table~\ref{tab:abla_ehc} shows that employing the proposed EHC module for initial LHFeat extraction achieves better speed and accuracy under different refinement configurations.
We also find that using multi-head self-attention for feature refinement provides a better speed-accuracy tradeoff. 
Finally, our overall architecture for extracting the LHFeat is considerably better than~\cite{SunHSC19}'s---the depth MAE is improved from $0.3002$ to $0.2835$ with FPS from $38$ to $52$.
All experiments in Table~\ref{tab:abla_ehc} deploy ResNet-50 as backbone and use the IDCT with $r=64$ for dense prediction.

\begin{table}
    \centering
    \begin{subtable}{\linewidth}
        \centering
        \begin{tabular}{c@{\hskip 4pt}c||ccc|c}
        \hline
        HC & Refine & MAE$\downarrow$ & RMSE$\downarrow$ & $\delta^1 \uparrow$ & FPS$\uparrow$ \\
        \hline\hline
        \cite{SunHSC19} & \multirow{2}{*}{-}       & 0.3090 & 0.5238 & 0.8158 & 49 \\
        EHC             &                          & \underline{0.3022} & \underline{0.5102} & \underline{0.8204} & 54 \\
        \hline
        \cite{SunHSC19} & \multirow{2}{*}{Bi-LSTM} & 0.3002 & 0.5147 & 0.8254 & 38 \\
        EHC             &                          & \underline{0.2928} & \underline{0.5036} & \underline{0.8294} & 41 \\
        \hline
        \cite{SunHSC19} & \multirow{2}{*}{MHSA} & 0.2915 & 0.5035 & 0.8331 & 47 \\
        EHC             &                       & \underline{\textbf{0.2835}} & \underline{\textbf{0.4916}} & \underline{\textbf{0.8389}} & 52 \\
        \hline
        \end{tabular}
        \caption{
            Comparison of the components for LHFeat extraction.
            The `HC' column indicates the height compression block, which produces the initial LHFeat from the backbone features.
            We compare the results of `no feature refinement', `refined by bidirectional LSTM'~\cite{HochreiterS97} (Bi-LSTM), and `refined by multi-head self-attention'~\cite{VaswaniSPUJGKP17} (MHSA). Refinement with MHSA achieves the most favorable results.
        }
        \label{tab:abla_ehc}
    \end{subtable}
    \begin{subtable}{\linewidth}
        \centering
        \begin{tabular}{cc||ccc|c}
        \hline
        $r$ & Basis & MAE$\downarrow$ & RMSE$\downarrow$ & $\delta^1 \uparrow$ & FPS$\uparrow$ \\
        \hline\hline
        \multirow{2}{*}{32}  & Interp. & 0.2886 & 0.5013 & 0.8356 & 52 \\ 
                             & IDCT    & \underline{0.2847} & \underline{0.4935} & \underline{0.8369} & 52 \\
        \hline
        \multirow{2}{*}{64}  & Interp. & 0.2880 & 0.4996 & 0.8351 & 52 \\ 
                             & IDCT    & \underline{\textbf{0.2835}} & \underline{\textbf{0.4916}} & \underline{0.8389} & 52 \\
        \hline
        \multirow{2}{*}{128} & Interp. & 0.2926 & 0.5043 & 0.8308 & 52 \\ 
                             & IDCT    & \underline{0.2850} & \underline{0.4955} & \underline{\textbf{0.8405}} & 52 \\
        \hline
        \multirow{2}{*}{256} & Interp. & 0.2937 & 0.5059 & 0.8260 & 52 \\ 
                             & IDCT    & \underline{0.2903} & \underline{0.5028} & \underline{0.8334} & 52 \\
        \hline
        \multirow{2}{*}{512} & Interp. & 0.3045 & 0.5189 & 0.8227 & 52 \\ 
                             & IDCT    & \underline{0.2913} & \underline{0.5040} & \underline{0.8341} & 52 \\
        \hline
        \end{tabular}
        \caption{
            Comparison on the different settings of the proposed horizon-to-dense module.
            The parameter $r$ denotes the number of components in a basis.
            We compare the two bases that implement the linear interpolation (Interp.) and the inverse discrete cosine transform (IDCT).
        }
        \label{tab:abla_basis}
    \end{subtable}
    \begin{subtable}{\linewidth}
        \centering
        \begin{tabular}{c||ccc|c}
        \hline
        Backbone & MAE$\downarrow$ & RMSE$\downarrow$ & $\delta^1 \uparrow$ & FPS$\uparrow$ \\
        \hline\hline
        ResNet34 & 0.2854 & 0.4976 & \textbf{0.8397} & \textbf{110}\\
        ResNet50 & \textbf{0.2835} & \textbf{0.4916} & 0.8389 & 52 \\
        \hline
        \end{tabular}
        \caption{
            Comparison of the results with different backbones.
        }
        \label{tab:abla_resnet}
    \end{subtable}

    \caption{
        Ablation study on depth modality using the \emph{ablation split} of  Matterport3D~\cite{ChangDFHNSSZZ17}.
        More details are in Sec.~\ref{ssec:exp_ablation}.
    }
    \label{tab:abla_h2d}
    \vspace{-2em}
\end{table}

\paragraph{Hyperparameters of horizon-to-dense.}
We compare the two operations---linear interpolation (spatial domain) and IDCT (frequency domain)---applied to dense prediction under different basis setups. 
As shown in Table~\ref{tab:abla_basis}, learning to predict in frequency domain (with IDCT) is consistently better than predicting in spatial domain (with linear interpolation) for dense depth estimation upon the compact LHFeat.
Interestingly, the number of components $r$ is not monotonic to the resulting accuracy, and we find $r=64$ is the best setting for our model.
As the compared operations introduce negligible computational cost, the FPSs are almost identical even if we increase $r$.
All experiments in Table~\ref{tab:abla_basis} share the same deep net setting that consists of ResNet-50, the proposed EHC, and the MHSA.

\paragraph{Comparison of the backbones.}
We compare the results of different backbones in Table~\ref{tab:abla_resnet}, where we find that employing ResNet-34 can almost double the FPS with only a little drop in accuracy comparing to ResNet-50.

\begin{table*}
    \centering
    \begin{tabular}{cc||ccccccc}
    \hline
    Dataset & Method & MRE & MAE & RMSE & RMSE (log) & $\delta^1$ & $\delta^2$ & $\delta^3$\\
    \hline\hline
    \multirow{7}{*}{Matterport3D} & FCRN~\cite{LainaRBTN16}            & 0.2409 & 0.4008 & 0.6704 & 0.1244 & 0.7703 & 0.9174 & 0.9617 \\
                                  & OmniDepth (bn)~\cite{ZioulisKZD18} & 0.2901 & 0.4838 & 0.7643 & 0.1450 & 0.6830 & 0.8794 & 0.9429 \\
                                  & Equi~\cite{WangYSCT20}             & 0.2074 & 0.3701 & 0.6536 & 0.1176 & 0.8302 & 0.9245 & 0.9577 \\
                                  & Cube~\cite{WangYSCT20}             & 0.2505 & 0.3929 & 0.6628 & 0.1281 & 0.7556 & 0.9135 & 0.9612 \\
                                  & BiFuse~\cite{WangYSCT20}           & 0.2048 & 0.3470 & 0.6259 & 0.1134 & 0.8452 & 0.9319 & 0.9632 \\
                                  & Ours                                & {\bf 0.1488} & {\bf 0.2862} & {\bf 0.5138} & {\bf 0.0871} & {\bf 0.8786} & {\bf 0.9519} & {\bf 0.9771}\\
    \hline\hline
    \multirow{7}{*}{Stanford2D3D} & FCRN~\cite{LainaRBTN16}            & 0.1837 & 0.3428 & 0.5774 & 0.1100 & 0.7230 & 0.9207 & 0.9731 \\
                                  & OmniDepth (bn)~\cite{ZioulisKZD18} & 0.1996 & 0.3743 & 0.6152 & 0.1212 & 0.6877 & 0.8891 & 0.9578 \\
                                  & Equi~\cite{WangYSCT20}             & 0.1428 & 0.2711 & 0.4637 & 0.0911 & 0.8261 & 0.9458 & 0.9800 \\
                                  & Cube~\cite{WangYSCT20}             & 0.1332 & 0.2588 & 0.4407 & 0.0844 & 0.8347 & 0.9523 & 0.9838 \\
                                  & BiFuse~\cite{WangYSCT20}           & 0.1209 & 0.2343 & 0.4142 & 0.0787 & 0.8660 & 0.9580 & 0.9860 \\
                                  & Ours                               & {\bf 0.1014} & {\bf 0.2027} & {\bf 0.3834} & {\bf 0.0668} & {\bf 0.9054} & {\bf 0.9693} & {\bf 0.9886} \\
    \hline
    \end{tabular}
    \caption{
        State-of-the-art comparison for depth estimation on real-world indoor 360 datasets---Matterport3D~\cite{ChangDFHNSSZZ17} and Stanford2D3D~\cite{ArmeniSZS17}.
        The evaluation protocol follows~\cite{WangYSCT20}, where the depth is clipped to 10 meter without depth median alignment.
    }
    \label{tab:sota_depth}
    \vspace{-1em}
\end{table*}

\subsection{Depth estimation}  \label{ssec:exp_depth}

\subsubsection{State-of-the-art comparison using the protocol of Wang~\etal~\cite{WangYSCT20}} \label{sssec:exp_depth_futen}
\paragraph{Datasets and evaluation protocol.}
We compare HoHoNet with state-of-the-art 360 depth estimation methods on real-world datasets following the testing protocol of \cite{WangYSCT20}.
Matterport3D~\cite{ChangDFHNSSZZ17} has 10{,}800 panoramas, and its training split contains 61 houses, and the testing results are reported on the merged official validation and test split.
Stanford2D3D~\cite{ArmeniSZS17} contains 1{,}413 panoramas, and the fold-1 is used where the fifth area is for testing, and the other areas are for training.
All the ERP images and depth maps are resized to $512 \times 1024$.
Standard depth estimation evaluation metrics---MRE, MAE, RMSE, RMSE (log), and $\delta$---are used.
Depths are clipped to 10 meters without median alignment.

\paragraph{Implementation details.}
We employ ResNet-50 as the backbone with the proposed EHC module for LHFeat extraction; the latent size $D$ of LHFeat is set to $256$; IDCT with $r=64$ components is applied to the model predictions.
We use Adam~\cite{KingmaB14} to optimize the L1 loss for $60$ epochs with a batch-size of $4$.
The learning rate is set to 1e-4, and we apply the polynomial learning rate decay with factor $0.9$.

\paragraph{Results.}
Table~\ref{tab:sota_depth} shows the comparisons with prior arts.
We demonstrate that the proposed HoHoNet outperforms the previous state-of-the-art, BiFuse~\cite{WangYSCT20}, by a large margin.
Note also that BiFuse takes both ERP and cubemap as their model inputs and thus requires two backbone networks.
HoHoNet has only one backbone and the compact LHFeat can achieve superior results, which shows the effectiveness of the proposed framework.

\begin{figure*}
    \centering
    \includegraphics[width=.9\linewidth]{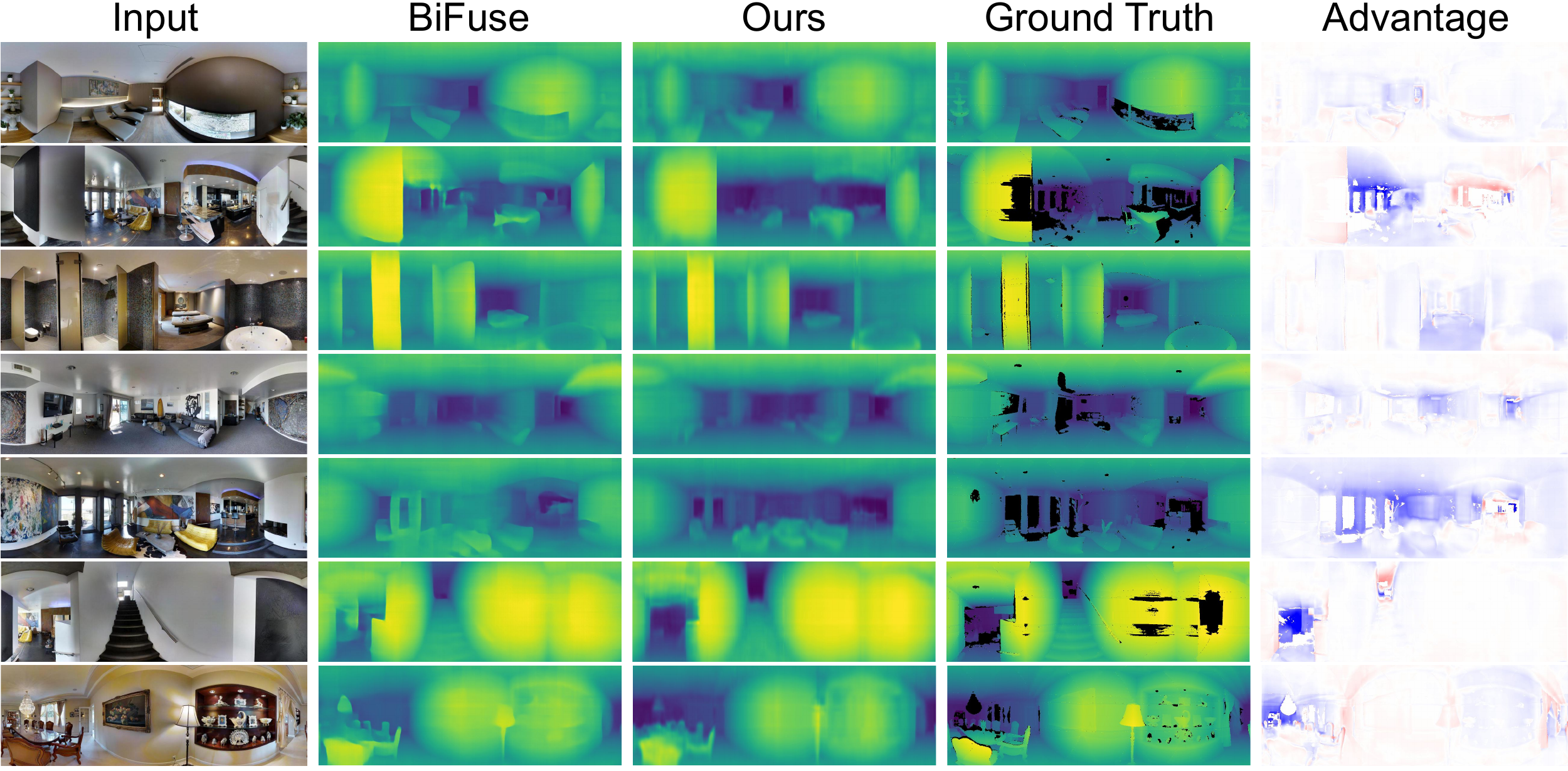}
    \caption{
        Qualitative comparison of the estimated dense depth with the prior art---BiFuse~\cite{WangYSCT20}.
        The `Advantage' column shows the MAE difference between ours and BiFuse's where the blue color indicates ours is better and the red color for vice versa.
        We find HoHoNet achieves good results in capturing the overall structure, but we also find some drawback in the visualization.
        First, HoHoNet's depth boundary is blurrier comparing to those of BiFuse.
        Second, some high-frequency signal in a column is discarded by HoHoNet.
        See the last row for an example.
        We find that \textit{i)} the boundary of the chairs in the left of the image is blurrier, and \textit{ii)} the lamp at the middle of the image is poorly reconstructed by HoHoNet while it seems to be easier to reconstruct from the conventional dense features.
        The intuitive reason for the qualitatively identified drawback is that the LHFeat focuses on learning the most prominent signals of a column, which makes it easier to optimize the training criterion.
    }
    \label{fig:depth_qual}
    \vspace{-1em}
\end{figure*}

A qualitative comparison with BiFuse~\cite{WangYSCT20} is provided in Fig.~\ref{fig:depth_qual}, where we download their code\footnote{\url{https://github.com/Yeh-yu-hsuan/BiFuse}} and the pre-trained weights for the comparison.
We find that HoHoNet is good at capturing the overall structure of the scene.
However, some drawbacks of HoHoNet are also observable through the visualization in Fig.~\ref{fig:depth_qual}.

\subsubsection{State-of-the-art comparison using the protocol of Jin~\etal~\cite{JinXZZTXYG20}} \label{sssec:exp_depth_jin}
\paragraph{Dataset and evaluation protocol.}
We also compare HoHoNet with another set of methods following the testing protocol of \cite{JinXZZTXYG20}.
A subset of the real-world Stanford2D3D~\cite{ArmeniSZS17} dataset with extra layout annotation is used, where there are only $404$ and $113$ panoramas for training and testing.
All the ERP images and depth maps are resized to $256 \times 512$.
Standard evaluation metrics---RMSE, MRE, log10, and $\delta^1$---for depth estimation are used.
Neither depth clipping nor median alignment is applied during evaluation.

\paragraph{Implementation details.}
The network and the training details are the same as in Sec.~\ref{sssec:exp_depth_futen}.
However, we find the training strategy of \cite{JinXZZTXYG20} is very different from ours.
For a fair comparison, we also report the results of training HoHoNet with the training protocol of \cite{JinXZZTXYG20}---SGD optimizer with a batch-size of $8$, learning rate of $0.01$, and weight decay set to 5e-4.

\paragraph{Results.}
The comparison on the Stanford2D3D subset is shown in Table~\ref{tab:s2d3d_small_depth}.
HoHoNet achieves the best accuracy under the same training protocol, and using Adam optimizer with our training setting can further improve the results.
Note that GeoReg360~\cite{JinXZZTXYG20} employs a ResNet-50 and a ResNet-34, and the network is jointly trained with the additional layout and semantic annotation.
Conversely, HoHoNet employs a single ResNet-50 and is only trained with depth modality, but still shows superior results, which further demonstrates the effectiveness of the proposed framework.

\begin{table}
\centering
\begin{subtable}{\linewidth}
    \centering
    \begin{tabular}{c||cccc}
    \hline
    Method & RMSE & MRE & log10 & $\delta^1$\\
    \hline\hline
    FCRN~\cite{LainaRBTN16}                        & 0.534 & 0.164 & 0.073 & 0.749\\
    UResNet~\cite{ZioulisKZD18}         & 0.590 & 0.187 & 0.084 & 0.711\\
    RectNet~\cite{ZioulisKZD18}         & 0.577 & 0.181 & 0.081 & 0.717\\
    Sph. FCRN~\cite{TatenoNT18}         & 0.523 & 0.145 & 0.067 & 0.783\\
    U-Net~\cite{RonnebergerFB15}                   & 0.472 & 0.140 & 0.062 & 0.803\\
    GeoReg360~\cite{JinXZZTXYG20}\textdagger       & 0.421 & 0.118 & 0.053 & 0.851\\
    Ours\textasteriskcentered                  & 0.408 & 0.111 & 0.050 & 0.867 \\
    Ours                 & {\bf 0.394} & {\bf 0.104} & {\bf 0.048} & {\bf 0.896} \\
    \hline
    \multicolumn{5}{l}{\textasteriskcentered\footnotesize{Using \cite{JinXZZTXYG20} training protocol for a fair comparison.}}\\
    \multicolumn{5}{l}{\textdagger\footnotesize{Using layout and semantic annotation.}}\\
    \end{tabular}
    \caption{
        Quantitative comparison for dense depth on Stanford2D3D~\cite{ArmeniSZS17} layout-available subset~\cite{ZouCSH18}.
        We strictly follow~\cite{JinXZZTXYG20} evaluation protocol.
        See detail in Sec.~\ref{sssec:exp_depth_jin}.
    }
    \label{tab:s2d3d_small_depth}
\end{subtable}
\begin{subtable}{\linewidth}
    \centering
    \begin{tabular}{c@{\hskip 4pt}c@{\hskip 4pt}c||c@{\hskip 4pt}c}
    \hline
    $H \times W$ & Input & Method & mIoU & mAcc \\
    \hline\hline
    \multicolumn{5}{l}{Simple backbone w/ low-resolution $360\degree$}\\
    \hline
    \multirow{5}{*}{$64 \times 128$}
      & RGB-D & Gauge Net~\cite{CohenWKW19} & 39.4 & 55.9\\ 
      & RGB-D & UGSCNN~\cite{JiangHKPMN19}  & 38.3 & 54.7\\
      & RGB-D & HexRUNet~\cite{ZhangLSC19}  & {\bf 43.3} & {\bf 58.6}\\
      & RGB-D & TangentImg~\cite{EderSLF20} & 37.5 & 50.2\\
      & RGB-D & Ours                        & 40.8 & 52.1\\
    \hline
    \multirow{2}{*}{$256 \times 512$}
      & RGB-D & TangentImg~\cite{EderSLF20} & 41.8 & {\bf 54.9}\\
      & RGB-D   & Ours                      & {\bf 43.3} & 53.9\\
    \hline
    \hline
    \multicolumn{5}{l}{ResNet backbone w/ high-resolution $360\degree$}\\
    \hline
    $2048 \times 4096$ & RGB   & TangentImg~\cite{EderSLF20} & 45.6 & {\bf 65.2}\\
    $1024 \times 2048$ & RGB   & Ours                        & {\bf 52.0} & 65.0\\
    \hline
    $2048 \times 4096$ & RGB-D & TangentImg~\cite{EderSLF20} & 51.9 & {\bf 69.1}\\
    $1024 \times 2048$ & RGB-D & Ours                        & {\bf 56.3} & 68.9\\
    \hline
    \end{tabular}

    \caption{
        Quantitative comparison for semantic segmentation on Stanford2D3D~\cite{ArmeniSZS17}.
        Results are averaged over the official 3 folds.
    }
    \label{tab:sem}
\end{subtable}
\begin{subtable}{\linewidth}
    \centering
    \begin{tabular}{c@{\hskip 3pt}c||cc|c}
    \hline
    \multirow{2}{*}{Method} & \multirow{2}{*}{Backbone} & \multicolumn{2}{c|}{IoU} & \multirow{2}{*}{FPS} \\
     &  & 3D & 2D & \\
    \hline\hline
    LayoutNet v2~\cite{ZouSPCSWCH19}          & ResNet-34 & 75.82 & 78.73 & 46 \\
    DuLa-Net v2~\cite{YangWPWSC19}            & ResNet-50 & 75.05 & 78.82 & 34 \\
    HorizonNet~\cite{SunHSC19}                & ResNet-50 & 79.11 & 81.71 & 31 \\
    AtlantaNet~\cite{PintoreAG20}             & ResNet-50 & {\bf 80.02} & 82.09 & 5 \\
    Ours & ResNet-34 & 79.88 & {\bf 82.32} & {\bf 110} \\
    \hline
    \end{tabular}

    \caption{
        Quantitative comparison for room layout estimation on MatterportLayout~\cite{ZouSPCSWCH19}.
    }
    \label{tab:layout}
\end{subtable}
\caption{State-of-the-art comparison on various datasets and different modalities.}
\vspace{-2em}
\end{table}

\subsection{Semantic segmentation}  \label{ssec:exp_semantic}
\paragraph{Dataset and evaluation protocol.}
We evaluate HoHoNet's semantic segmentation performance on Stanford2D3D~\cite{ArmeniSZS17} dataset.
As previous work, we report the averaged results from the official 3-fold cross-validation splits, using standard semantic segmentation evaluation metrics---class-wise mIoU and class-wise mAcc.

\paragraph{Implementation detail.}
The architecture setting of HoHoNet for semantic segmentation is almost the same as for depth estimation in Sec.~\ref{sssec:exp_depth_futen} except the last layer has $E=Nr=13\cdot64=832$ channels.
To compare with methods using a simple backbone under low resolution, we follow~\cite{EderSLF20,JiangHKPMN19,ZhangLSC19} to construct a shallow U-Net but purely with planar CNN.
For results on high resolution, we use ResNet-101 as backbone.
We use Adam~\cite{KingmaB14} to optimize the cross-entropy loss for $60$ epochs with a batch-size of $4$.
The learning rate is 1e-4 with polynomial decay of factor $0.9$.

\paragraph{Results.}
Table~\ref{tab:sem} shows the comparison with previous methods.
On the lowest resolution, HexRUNet~\cite{ZhangLSC19}, with a specially designed kernel on icosahedron representation, achieves the best result.
Ours with purely planar CNNs and compact LHFeat is still competitive with the distortion mitigated methods under the low-resolution settings.
When scaling to a high resolution, we achieve similar mACC with the recent state-of-the-art~\cite{EderSLF20}, while our mIoU is significantly better.
Note that the results of \cite{EderSLF20} are obtained from a stronger FCN-ResNet101 backbone and a higher input resolution.
Limited by our device and ERP projection, we can only train on a lower $1024 \times 2048$ resolution but still obtain competitive performance with the current state-of-the-art on $360\degree$ semantic segmentation.

\subsection{Room layout estimation} \label{ssec:exp_layout}
\paragraph{Dataset and evaluation protocol.}
We use MatterportLayout~\cite{ZouSPCSWCH19,WangYSCT20Layout} dataset, which is a real-world 360 Manhattan layout dataset.
The official evaluation function\footnote{\label{urllayoutnetv2}\url{https://github.com/zouchuhang/LayoutNetv2}} for 2D IoU and 3D IoU is used directly, where the 2D IoU is measured by projecting floor corners to an aligned floor, while 3D IoU is for pop-up view considering both floor and ceiling corners.

\paragraph{Implementation details.}
HoHoNet is compatible with the 1D layout representation proposed by HorizonNet~\cite{SunHSC19}.
Since our main focus is not to design a new method for layout reconstruction, we use~\cite{SunHSC19}'s loss, training protocol, and post-processing algorithm directly.
We find HoHoNet with ResNet-34 shows slightly better accuracy than ResNet-50 in validation, so we use the simpler ResNet-34 as backbone.

\paragraph{Results.}
The comparison with previous methods on MatterportLayout is shown in Table~\ref{tab:layout}.
The FPSs are obtained using the official codes\footnoteref{urllayoutnetv2}\footnote{\url{https://github.com/SunDaDenny/DuLa-Net}}\footnote{\url{https://github.com/sunset1995/HorizonNet}}\footnote{\label{urlatlantanet}\url{https://github.com/crs4/AtlantaNet}} and measured by the averaged feed-forward times of the models on a GeForce RTX 2080 Ti.
The result of AtlantaNet~\cite{PintoreAG20} is obtained from their official new pre-trained weights\footnoteref{urlatlantanet} with aligned data split and re-evaluated by the official evaluation function\footnoteref{urllayoutnetv2}.
Our result is on par with the state-of-the-art AtlantaNet but $22\times$ faster.
HoHoNet also outperforms HorizonNet~\cite{SunHSC19} by $+0.77$ 3D IoU and $+0.61$ 2D IoU, and is $3.5\times$ faster, which shows the effectiveness of the designed architecture.

\subsection{Results on non-gravity-aligned views} \label{ssec:poor_pose}

In Fig.~\ref{fig:depth_motivation}, we show that the structure signals of an image column suffer more losses in compression if the image's $y$-axis is not aligned with the gravity.
Though the 360 data in all benchmarks we use are mostly well-aligned with gravity, the captured $360\degree$ views could be non-gravity-aligned in practice.
In Table~\ref{tab:poor_pose}, we show the vulnerability of our model to heavy pitch or roll rotation (see Fig.~\ref{fig:pitch_rotation} and Fig.~\ref{fig:roll_rotation} for visualization).
The pre-trained model in our ablation study takes the rotated images directly as input, and the output depth maps are rotated back to the original view for a fair comparison.
As expected, the pre-trained model performs poorly when input $360\degree$ views are not gravity-aligned.
Introducing $10\degree$ of pitch or roll rotation increases MAE from $28.45$cm to more than $44$cm.
A simple solution is to use the IMU sensor or 360 vanishing point detection algorithm~\cite{ZhangSTX14,ZouCSH18} to ensure gravity alignment (the VP alignment is also a standard step in 360 layout benchmark~\cite{WangYSCT20Layout,ZouCSH18,ZouSPCSWCH19}).

We also show the results by training with $\mathcal{U}(-30\degree, 30\degree)$ pitch/roll rotation as data augmentation, which makes the model much more robust against the non-canonical view but sacrifices the test-time performance when input $360\degree$ view are  gravity-aligned (MAE$\uparrow$ from $28.35$cm to $30.92$cm).

\begin{table}
    \centering
    \begin{tabular}{c||c|c@{\hskip 7pt}c@{\hskip 7pt}c@{\hskip 7pt}c}
    \hline
    \multirow{2}{*}{\makecell{Training\\Rot. Aug.}} & \multirow{2}{*}{\makecell{Testing\\Cam. Rot.}} & \multicolumn{4}{c}{MAE (cm)} \\
     & & $0\degree$ & $10\degree$ & $20\degree$ & $30\degree$ \\
    \hline\hline
               & \multirow{2}{*}{Pitch} & {\bf 28.35} & 44.88 & 62.77 & 75.79 \\
    \checkmark & & 30.92 & {\bf 31.30} & {\bf 31.80} & {\bf 32.97} \\
    \hline
               & \multirow{2}{*}{Roll} & {\bf 28.35} & 44.32 & 61.90 & 75.11 \\
    \checkmark & & 30.92 & {\bf 31.32} & {\bf 31.80} & {\bf 32.90} \\
    \hline
    \end{tabular}

    \caption{
        Vulnerability to non-gravity-aligned views.
    }
    \label{tab:poor_pose}
    \vspace{-1em}
\end{table}

\section{Conclusion}
This work presents a novel framework, HoHoNet, which is the first step to learning compact latent horizontal features for dense modalities modeling of omnidirectional images.
HoHoNet is fast, versatile, and accurate for solving layout reconstruction, depth estimation, and semantic segmentation with accuracy on par with or better than the state-of-the-art.

\paragraph{Acknowledgements:}
This work was supported in part by the MOST, Taiwan under Grants 110-2634-F-001-009 and 110-2634-F-007-016, MOST Joint Research Center for AI Technology and All Vista Healthcare. We thank National Center for High-performance Computing (NCHC) for providing computational and storage resources.

\clearpage
\onecolumn
\setcounter{section}{0}
\renewcommand\thesection{\Alph{section}}

\noindent{\bf\Large Supplementary material}

\section{Network architecture diagram}
\begin{figure*}[h]
    \centering
    \includegraphics[width=0.9\linewidth]{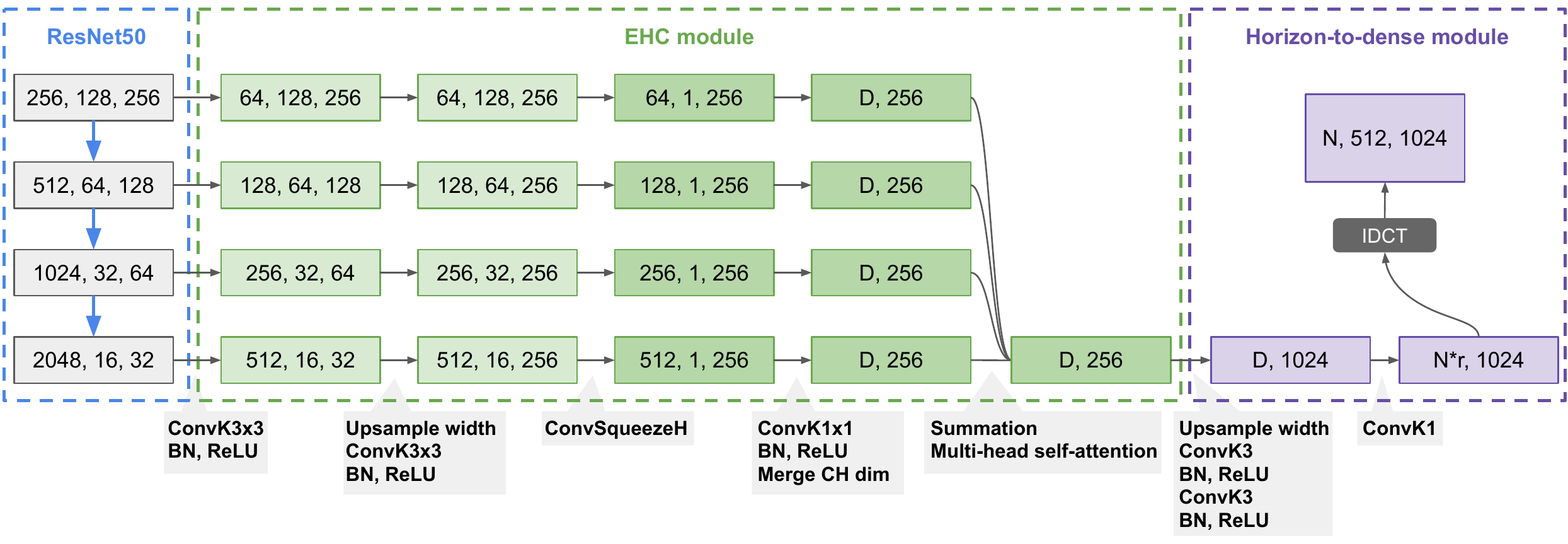}
    \caption{
        The detailed network architecture with ResNet50~\cite{HeZRS16} backbone.
    }
    \label{fig:detail_arch}
\end{figure*}

We show the detailed architecture diagram in Fig.~\ref{fig:detail_arch}.
The shape of each feature tensor is denoted as ``\emph{\# of channels, height, width}'' within the box.
The height and width of the input panorama are assumed to be $512$ and $1024$ respectively.
$D$ and $E$ are hyperparameters.
The $\operatorname{ConvSqueezeH}$ layer is a depthwise convolution layer with kernel size set to the prior known input feature height without padding, which produces output feature height $1$.

\section{Comparing EHC block and HC block~\cite{SunHSC19}}
\begin{figure}[h]
    \centering
    \begin{subfigure}[t]{0.30\linewidth}
      \centering
      \includegraphics[height=1.3\linewidth]{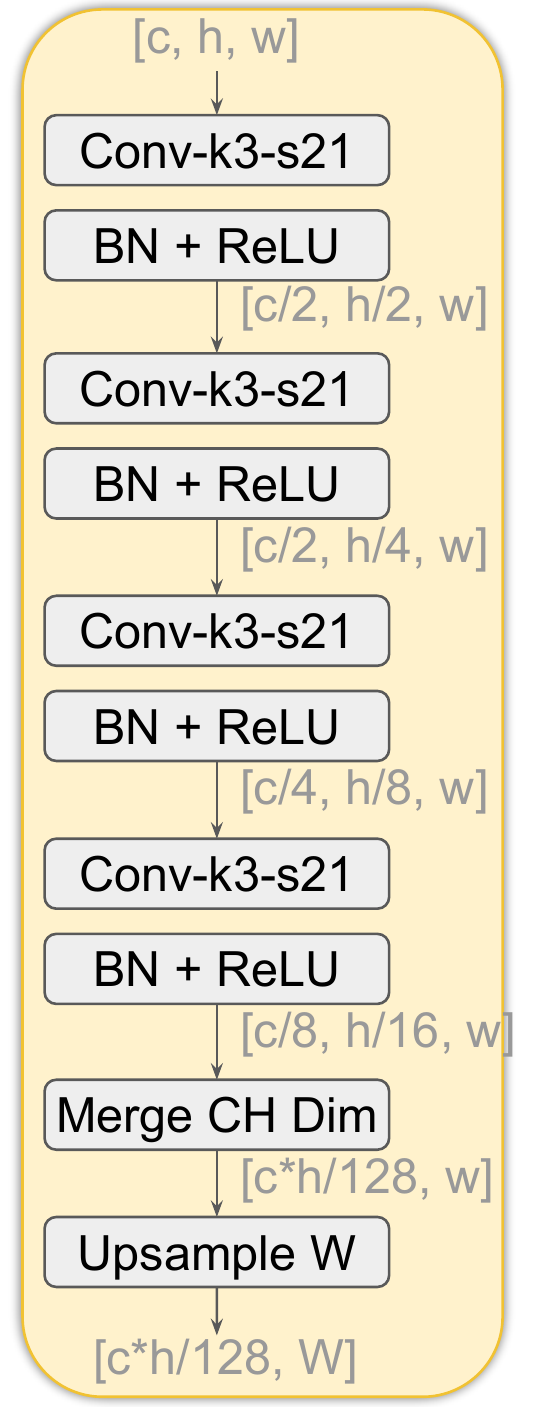}
      \caption{The HC block in~\cite{SunHSC19}.}
    \end{subfigure}
    \begin{subfigure}[t]{0.30\linewidth}
      \centering
      \includegraphics[height=1.3\linewidth]{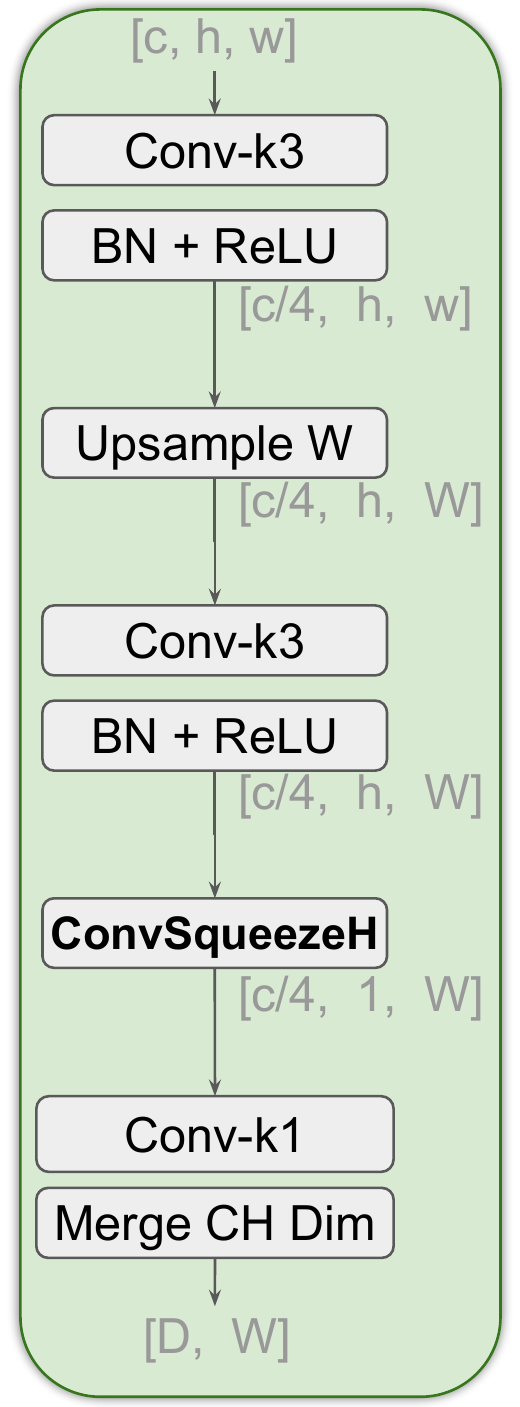}
      \caption{The proposed EHC block.}
    \end{subfigure}
    \caption{Comparison of the proposed EHC block and the HC block in~\cite{SunHSC19}.}
    \label{fig:detail_ehc_hc_comparison}
\end{figure}

The height compression block aims to squeeze a 2D feature from the backbone to produce a 1D horizontal feature.
Fig.~\ref{fig:detail_ehc_hc_comparison} shows the architecture of our \emph{Efficient Height Compression block} (EHC block) and the one of HC block~\cite{SunHSC19} for comparison.
The HC block \cite{SunHSC19} employs a sequence of convolution layers to gradually reduce the number of channels and heights, while we first use a convolution layer for channel reduction and then use bilinear upsampling and $\operatorname{ConvSqueezeH}$ layer to produce the features in horizontal shape.
We show in our ablation experiments that replacing the HC block~\cite{SunHSC19} with the proposed ECH block leads to better speed and accuracy.

\section{Detailed semantic segmentation results}
We show detailed per-class IoU and per-class Acc for semantic segmentation in Table~\ref{tab:detail_percls}.
We achieve the best IoU on 10 out of 13 classes and superior overall mIoU; we achieve best Acc on 7 out of 13 classes and comparable overall mAcc.
\begin{table*}[h]
\begin{subtable}{\linewidth}
    \centering
    \begin{tabular}{@{\hskip 3pt}c@{\hskip 3pt}|@{\hskip 3pt}c@{\hskip 3pt}|@{\hskip 3pt}c@{\hskip 6pt}c@{\hskip 6pt}c@{\hskip 6pt}c@{\hskip 6pt}c@{\hskip 6pt}c@{\hskip 6pt}c@{\hskip 6pt}c@{\hskip 6pt}c@{\hskip 6pt}c@{\hskip 6pt}c@{\hskip 6pt}c@{\hskip 6pt}c@{\hskip 3pt}}
    \hline
    Method & overall & beam & board & bookcase & ceiling & chair & clutter & column & door & floor & sofa & table & wall & window\\
    \hline\hline
    \multicolumn{15}{c}{Low-resolution RGB-D}\\
    \hline
    UGSCNN~\cite{JiangHKPMN19}  & 38.3 &  8.7 & 32.7 & 33.4 & 82.2 & 42.0 & 25.6 & 10.1 & 41.6 & 87.0 &  7.6 & 41.7 & 61.7 & 23.5\\
    HexRUNet~\cite{ZhangLSC19}  & 43.3 & {\bf 10.9} & 39.7 & 37.2 & 84.8 & 50.5 & 29.2 & 11.5 & 45.3 & 92.9 & 19.1 & 49.1 & 63.8 & 29.4\\
    TangentImg~\cite{EderSLF20} & 37.5 & {\bf 10.9} & 26.6 & 31.9 & 82.0 & 38.5 & 29.3 &  5.9 & 36.2 & 89.4 & 12.6 & 40.4 & 56.5 & 26.7\\
    Ours                        & 40.8 &  3.6 & 43.5 & 40.6 & 81.8 & 41.3 & 27.7 &  9.2 & 52.0 & 92.2 &  9.4 & 44.6 & 61.6 & 23.4\\
    \hline\hline
    \multicolumn{15}{c}{High-resolution RGB-D}\\
    \hline
    TangentImg~\cite{EderSLF20} & 51.9 &  4.5 & 49.9 & 50.3 & 85.5 & {\bf 71.5} & 42.4 & 11.7 & 50.0 & 94.3 & 32.1 & {\bf 61.4} & 70.5 & 50.0\\
    Ours                        & {\bf 56.3} &  7.4 & {\bf 62.3} & {\bf 55.5} & {\bf 87.0} & 66.4 & {\bf 44.3} & {\bf 19.2} & {\bf 66.5} & {\bf 96.1} & {\bf 43.3} & 60.1 & {\bf 72.9} & {\bf 51.4}\\
    \hline
    \end{tabular}
    \caption{
    Per-class IoU (\%).
    }
\end{subtable}
\par\medskip
\begin{subtable}{\linewidth}
    \centering
    \begin{tabular}{@{\hskip 3pt}c@{\hskip 3pt}|@{\hskip 3pt}c@{\hskip 3pt}|@{\hskip 3pt}c@{\hskip 6pt}c@{\hskip 6pt}c@{\hskip 6pt}c@{\hskip 6pt}c@{\hskip 6pt}c@{\hskip 6pt}c@{\hskip 6pt}c@{\hskip 6pt}c@{\hskip 6pt}c@{\hskip 6pt}c@{\hskip 6pt}c@{\hskip 6pt}c@{\hskip 3pt}}
    \hline
    Method & overall & beam & board & bookcase & ceiling & chair & clutter & column & door & floor & sofa & table & wall & window\\
    \hline\hline
    \multicolumn{15}{c}{Low-resolution RGB-D}\\
    \hline
    UGSCNN~\cite{JiangHKPMN19}  & 54.7 & 19.6 & 48.6 & 49.6 & 93.6 & 63.8 & 43.1 & 28.0 & 63.2 & 96.4 & 21.0 & 70.0 & 74.6 & 39.0\\
    HexRUNet~\cite{ZhangLSC19}  & 58.6 & 23.2 & 56.5 & 62.1 & 94.6 & 66.7 & 41.5 & 18.3 & 64.5 & 96.2 & 41.1 & {\bf 79.7} & 77.2 & 41.1\\
    TangentImg~\cite{EderSLF20} & 50.2 & {\bf 25.6} & 33.6 & 44.3 & 87.6 & 51.5 & 44.6 & 12.1 & 64.6 & 93.6 & 26.2 & 47.2 & 78.7 & 42.7\\
    Ours                        & 52.1 &  9.5 & 56.5 & 56.6 & 95.1 & 57.9 & 40.7 & 12.5 & 64.5 & 96.8 & 10.6 & 69.1 & 79.3 & 28.4\\
    \hline\hline
    \multicolumn{15}{c}{High-resolution RGB-D}\\
    \hline
    TangentImg~\cite{EderSLF20} & {\bf 69.1} & 22.6 & 62.0 & 70.0 & 90.3 & {\bf 84.7} & 55.5 & {\bf 41.4} & 76.7 & 96.9 & {\bf 70.3} & 73.9 & 80.1 & {\bf 74.3}\\
    Ours                        & 68.9 & 16.7 & {\bf 79.0} & {\bf 71.8} & {\bf 96.4} & 79.2 & {\bf 59.7} & 26.9 & {\bf 77.7} & {\bf 98.2} & 58.0 & 79.6 & {\bf 85.9} & 66.3\\
    \hline
    \end{tabular}
    \caption{
    Per-class Acc (\%).
    }
\end{subtable}
\caption{
Detailed quantitative per-class results on Stanford2D3D~\cite{ArmeniSZS17} with RGB-D as input.
}
\label{tab:detail_percls}
\end{table*}

\section{Detailed layout estimation results}
We show detailed quantitative results for room layout under different numbers of ground truth 2D corners in Table~\ref{tab:detail_layout}.
Our training protocol and layout formalization are identical to HorizonNet~\cite{SunHSC19}, while we observe improvements (except rooms with six corners) by using our network architecture.
In comparison with the most recent state-of-the-art---AtlantaNet~\cite{PintoreAG20}, we show better results on scenes with fewer corners and similar accuracy on overall scenes; meanwhile, our model is $22\times$ faster than AtlantaNet~\cite{PintoreAG20}.
\begin{table}[h]
    \centering
    \begin{tabular}{c|c|c|cccc}
    \hline
    \multirow{2}{*}{Method} & \multirow{2}{*}{Metric} & \multicolumn{5}{c}{\# of corners}  \\
    \cline{3-7}
     & & overall & 4 & 6 & 8 & 10+ \\
    \hline\hline
    LayoutNet v2~\cite{ZouSPCSWCH19}          & \multirow{5}{*}{{\bf 3D IoU (\%)}} & 75.82 & 81.35 & 72.33 & 67.45 & 63.00 \\
    DuLa-Net v2~\cite{YangWPWSC19}            &                                    & 75.07 & 77.02 & 78.79 & 71.03 & 63.27 \\
    HorizonNet~\cite{SunHSC19}                &                                    & 79.11 & 81.88 & {\bf 82.26} & 71.78 & 68.32 \\
    AtlantaNet~\cite{PintoreAG20}             &                                    & {\bf 80.02} & 82.09 & 82.08 & {\bf 75.19} & {\bf 71.61} \\
    Ours                                      &                                    & 79.88 & {\bf 82.64} & 82.16 & 73.65 & 69.26 \\
    \hline\hline
    LayoutNet v2~\cite{ZouSPCSWCH19}          & \multirow{5}{*}{{\bf 2D IoU (\%)}} & 78.73 & 84.61 & 75.02 & 69.79 & 65.14 \\
    DuLa-Net v2~\cite{YangWPWSC19}            &                                    & 78.82 & 81.12 & 82.69 & 74.00 & 66.12 \\
    HorizonNet~\cite{SunHSC19}                &                                    & 81.71 & 84.67 & {\bf 84.82} & 73.91 & 70.58 \\
    AtlantaNet~\cite{PintoreAG20}             &                                    & 82.09 & 84.42 & 83.85 & {\bf 76.97} & {\bf 73.18} \\
    Ours                                      &                                    & {\bf 82.32} & {\bf 85.26} & 84.81 & 75.59 & 70.98 \\
    \hline
    \end{tabular}
    \caption{
        Detailed quantitative comparison for room layout estimation on MatterportLayout~\cite{ZouSPCSWCH19} under different numbers of ground-truth corners.
    }
    \label{tab:detail_layout}
    \vspace{-1em}
\end{table}

\section{More qualitative comparisons for depth estimation}
\begin{figure*}[h]
    \vspace{-1em}
    \centering
    \includegraphics[width=\linewidth]{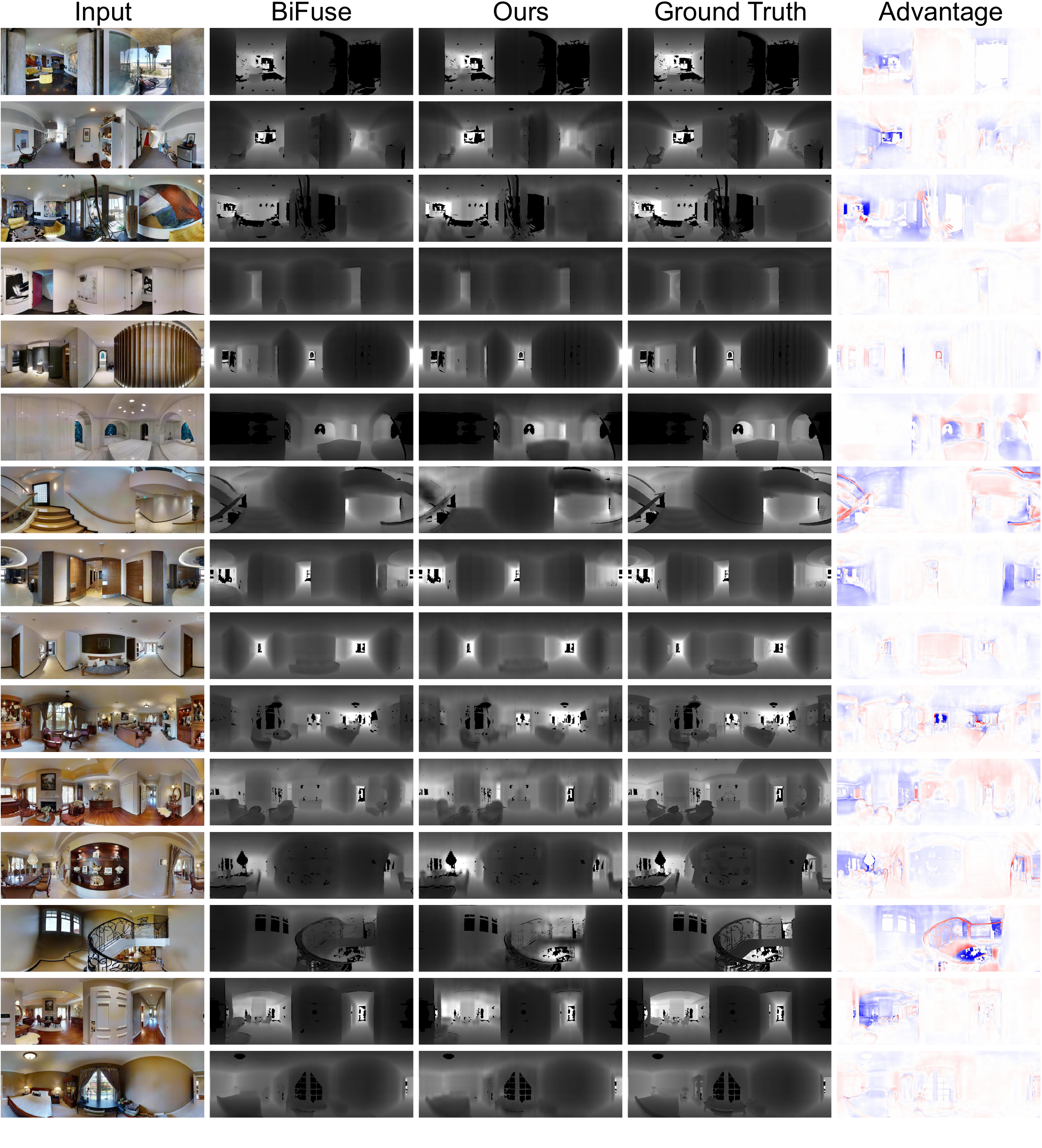}
    \caption{
    More qualitative comparisons of the estimated dense depth with the prior art---BiFuse~\cite{WangYSCT20}.
    The `Advantage' column shows the MAE difference between ours and BiFuse's where the blue color indicates ours is better and the red color for vice versa.
    }
    \label{fig:more_depth_compare}
\end{figure*}

We show more qualitative comparisons with the prior art---BiFuse~\cite{WangYSCT20}---in Fig.~\ref{fig:more_depth_compare}.
BiFuse's results are obtained from their official released model trained on the real-world Matterport3D~\cite{ChangDFHNSSZZ17} dataset.

\section{Qualitative results for semantic segmentation}
\begin{figure*}[h]
    \vspace{-1em}
    \centering
    \includegraphics[width=0.8\linewidth]{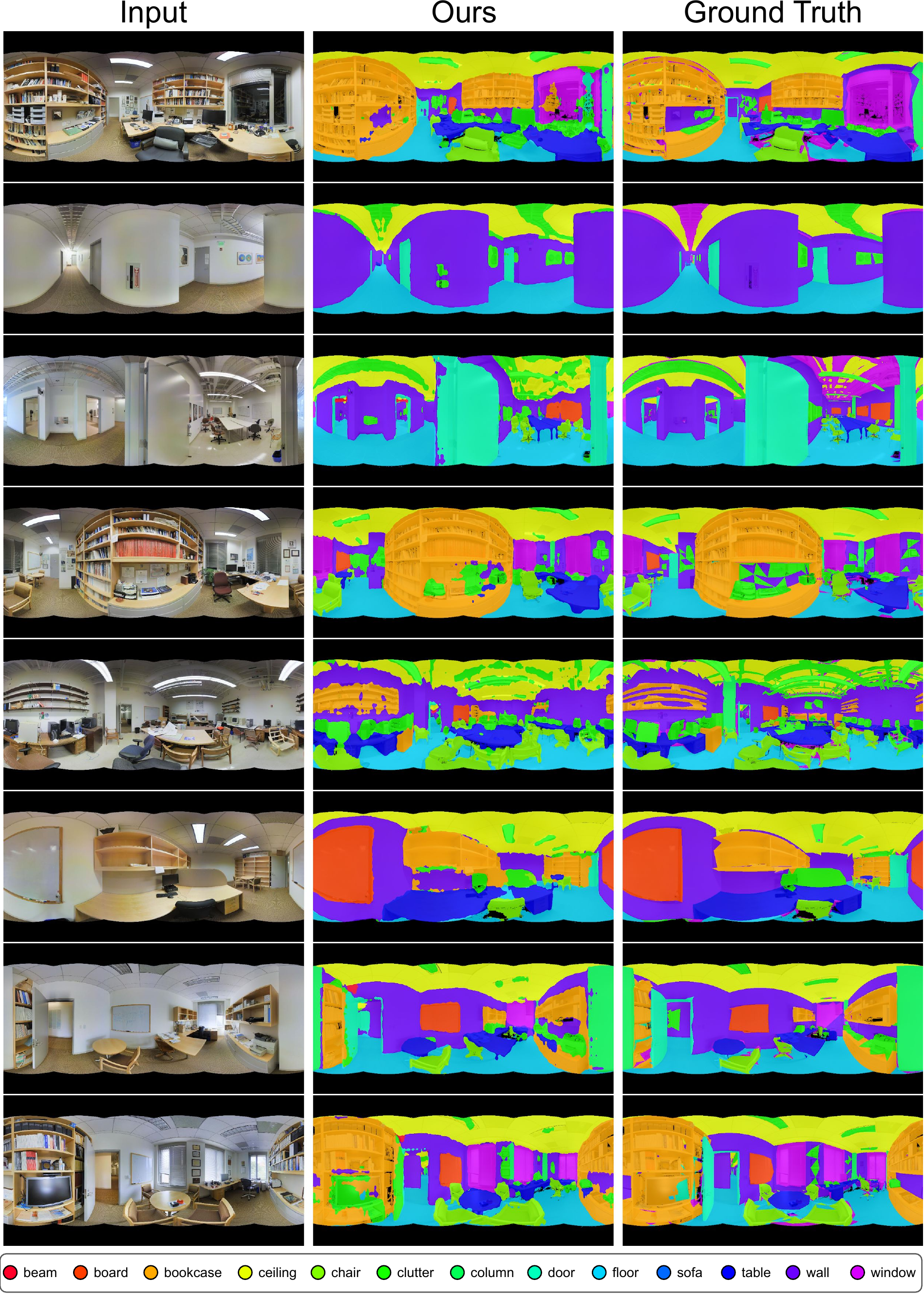}
    \caption{
    Qualitative results for semantic segmentation on Stanford2D3D~\cite{ArmeniSZS17} dataset.
    }
    \label{fig:sem_qual}
\end{figure*}

Qualitative results for semantic segmentation on Stanford2D3D~\cite{ArmeniSZS17} dataset are shown in Fig.~\ref{fig:sem_qual}.
We fail to build the prior art~\cite{EderSLF20} from their public release for semantic segmentation on high-resolution panorama, so we only show our results.

\section{Qualitative comparisons for layout estimation}
\begin{figure*}[h]
    \centering
    \includegraphics[width=\linewidth]{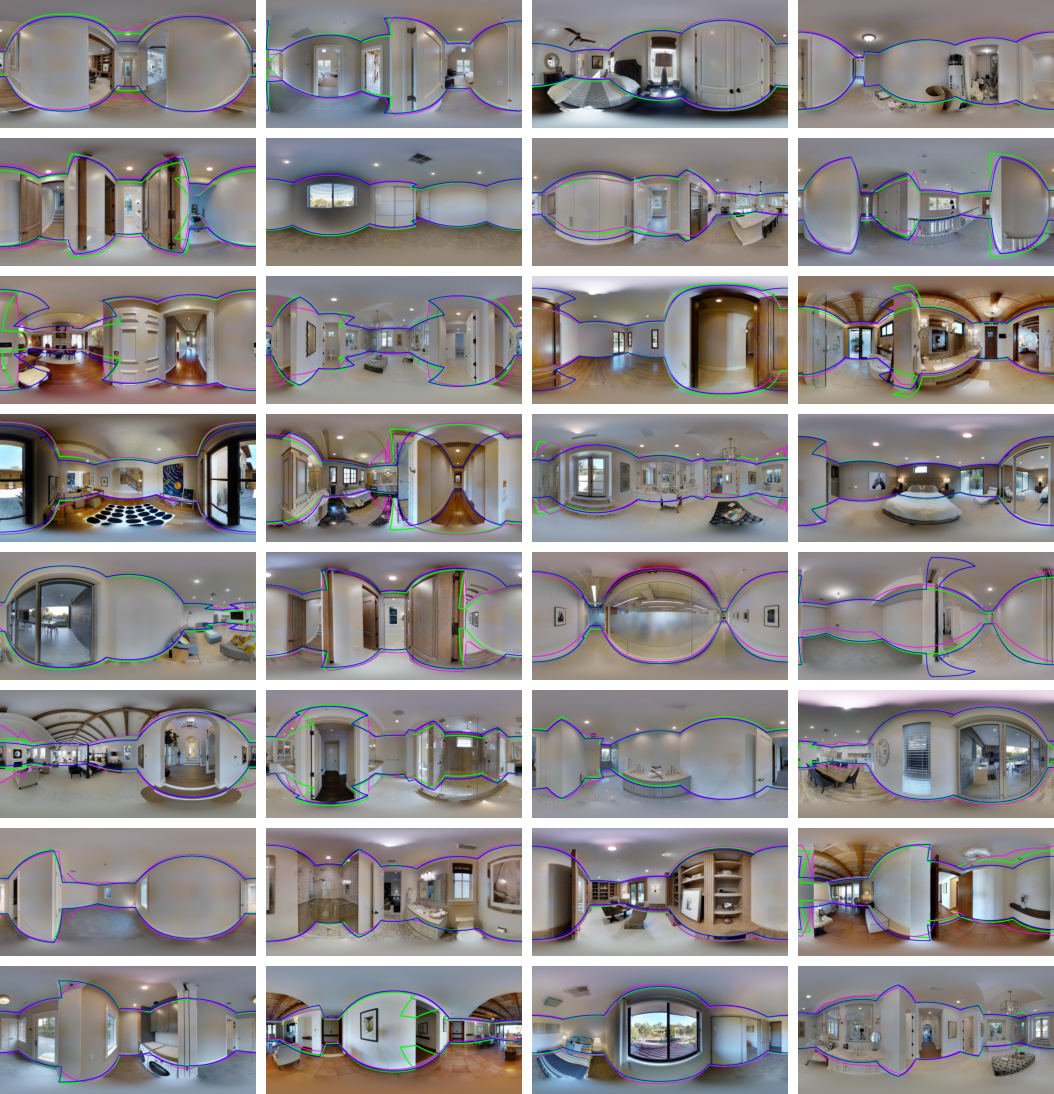}
    \caption{
    Qualitative comparisons for room layout estimated with the competitive AtlantaNet~\cite{PintoreAG20}.
    The \textcolor{green}{green}, \textcolor{magenta}{magenta}, and \textcolor{blue}{blue} are the ground truth layout, AtlantaNet's results, and our results respectively.
    }
    \label{fig:layout_qual}
\end{figure*}

We show qualitative comparisons for room layout estimation with the prior art---AtlantaNet~\cite{PintoreAG20}---in Fig.~\ref{fig:layout_qual}.
The results of AtlantaNet are obtained from their official code and pre-trained weights.
We use ~\cite{SunHSC19} post-processing algorithm to produce Manhattan layouts; AtlantaNet~\cite{PintoreAG20}'s algorithm generates less restrictive Atlanta layouts.
Our model achieves promising results comparable to the most recent AtlantaNet~\cite{PintoreAG20}, while our model runs $22\times$ faster.

\twocolumn
{\small
\bibliographystyle{ieee_fullname}
\bibliography{egbib}
}

\end{document}